\pdfoutput=1
\documentclass[11pt]{article}

\usepackage{amsmath}
\usepackage[ruled,vlined]{algorithm2e}

\usepackage[most]{tcolorbox}
\tcbuselibrary{breakable}
\usepackage{fontawesome5}

\usepackage{xcolor}
\usepackage[table]{xcolor}
\usepackage{makecell}

\usepackage[final]{acl}

\usepackage{multirow}
\usepackage{graphicx} 
\usepackage{booktabs}
\usepackage{subcaption}

\usepackage{times}
\usepackage{latexsym}

\usepackage[T1]{fontenc}
\usepackage[utf8]{inputenc}

\usepackage{microtype}

\usepackage{inconsolata}

\usepackage{graphicx}
\usepackage{enumitem}

\title{Enhancing Multi-Agent Debate System Performance via Confidence Expression}

\author{
  Zijie Lin, Bryan Hooi \\
  National University of Singapore \\
  \texttt{lin.zijie@u.nus.edu} \quad \texttt{bhooi@comp.nus.edu.sg}
}

\begin{document}
\maketitle
\begin{abstract}
Generative Large Language Models (LLMs) have demonstrated remarkable performance across a wide range of tasks. Recent research has introduced Multi-Agent Debate (MAD) systems, which leverage multiple LLMs to simulate human debate and thereby improve task performance. However, while some LLMs may possess superior knowledge or reasoning capabilities for specific tasks, they often struggle to clearly communicate this advantage during debates, in part due to a lack of confidence expression. Moreover, inappropriate confidence expression can cause agents in MAD systems to either stubbornly maintain incorrect beliefs or converge prematurely on suboptimal answers, ultimately reducing debate effectiveness and overall system performance. To address these challenges, we propose incorporating confidence expression into MAD systems to allow LLMs to explicitly communicate their confidence levels. To validate this approach, we develop \textbf{ConfMAD}, a MAD framework that integrates confidence expression throughout the debate process. Experimental results demonstrate the effectiveness of our method, and we further analyze how confidence influences debate dynamics, offering insights into the design of confidence-aware MAD systems.

\end{abstract}

\section{Introduction}

Multiple studies have demonstrated that LLMs possess emergent reasoning and reflection capabilities \citep{wang2022self, wei2022chain, madaan2023self}. By effectively harnessing these abilities, researchers can enhance the accuracy of LLM responses, minimize hallucinations, and strengthen reasoning capabilities. Building upon this foundation, \citet{du2023improving}, drawing inspiration from The Society of Mind \cite{minsky1986society} and multi-agent frameworks, proposed utilizing multi-agent systems composed of multiple LLMs to achieve superior performance on various tasks. Specifically, they developed a Multi-Agent Debate (MAD) system where, when presented with a query, multiple instances of LLM agents first generate independent candidate responses. Subsequently, these agents engage in a structured debate about these responses, iteratively refining and updating their own contributions throughout the process.

\begin{figure}[t]
  \includegraphics[width=\columnwidth]{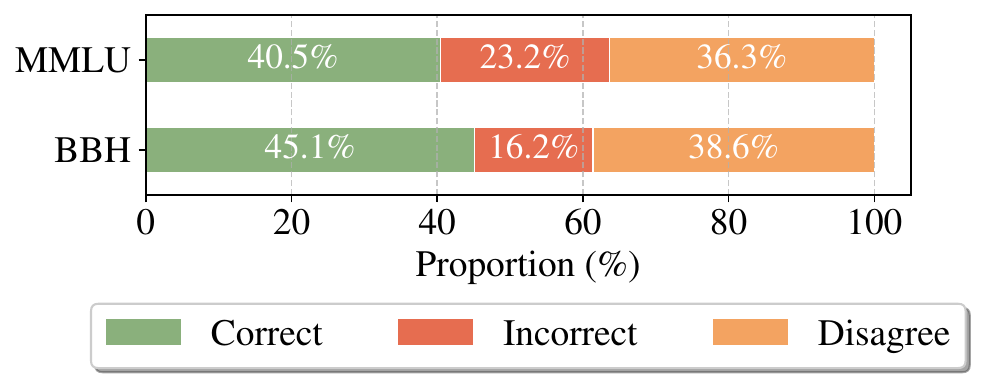}
  \caption{Debate outcomes in the initial round when only one LLM is initially correct. “Correct” indicates convergence to the right answer, “Incorrect” to the wrong one, and “Disagree” means no consensus. Debaters are GPT-4o-mini and Phi-4.}
  \label{fig:BarChart}
\end{figure}

However, we've identified a key issue with current MAD systems. LLM agents with different knowledge and capabilities don't explicitly express their confidence level regarding their arguments and knowledge during communication, preventing full utilization of each agent's relative strengths. This issue may limit the performance of current MAD systems, potentially leading to failure to converge or even convergence towards incorrect answers. As shown in Figure~\ref{fig:BarChart}, in a basic debate setting \citep{du2023improving}, when only one LLM provided the correct answer in the initial round on the BBH and MMLU datasets, fewer than 50\% of such cases ultimately converged to the correct answer.

\begin{figure}[t]
  \includegraphics[width=\columnwidth]{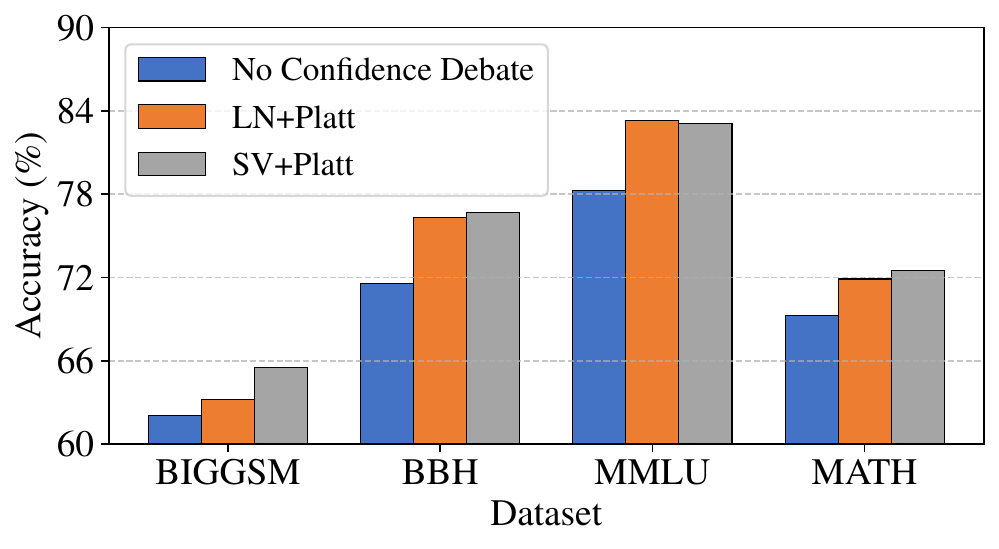}
  \caption{Accuracy of debates with and without confidence across different benchmarks using GPT-4o-mini and LLaMA-3.1-70B-Instruct. LN+Platt and SV+Platt indicate different settings of debating with confidence.}
  \label{fig:AccPlot1}
\end{figure}

To address the above issue, we propose an intuitive solution by incorporating confidence expression for each LLM instance in MAD systems. During the debate process, for a given query, each LLM agent's debate content includes not only arguments (reasons) and an answer, but also a confidence score. Additionally, considering that modern deep neural networks often exhibit overconfidence \citep{nguyen2015deep, guo2017calibration, kadavath2022language, mielke2022reducing, xiong2023can}, we introduce calibration methods to further investigate the impact of confidence on the MAD system. Based on these considerations, we developed a MAD framework with confidence expression called \textbf{ConfMAD}\footnote{Code is at \url{https://github.com/Enqurance/ConfMAD}}. We evaluated ConfMAD on various benchmarks. As shown in Figure \ref{fig:AccPlot1}, our results demonstrate that introducing confidence expression can lead to notable accuracy improvements across multiple benchmarks. Building on this foundation, we provide detailed discussions on how confidence scores influence the MAD debate process and offer some insights for developing better confidence-aware MAD systems.

In summary, the contributions of our work are summarized as follows:

\begin{itemize}[leftmargin=*]
  \item We propose ConfMAD, a MAD system that incorporates different confidence scores and calibration methods to enhance debate performance by enabling LLM agents to express confidence during interactions.
  \item We evaluate ConfMAD across multiple benchmarks. Results show that the confidence expression mechanism effectively improved the performance of MAD systems.
  \item We explore the contribution of confidence scores to MAD system performance. Our findings indicate that confidence scores not only improve individual LLM accuracy but also enhance the system’s ability to reach correct consensus through more effective agent interactions. Furthermore, we provide an in-depth analysis of how different confidence expression and calibration methods influence debate dynamics, offering insights for designing more robust confidence-aware MAD systems.
\end{itemize}

\section{Debate Framework}

\begin{figure*}[t]
  \centering
  \includegraphics[width=\textwidth]{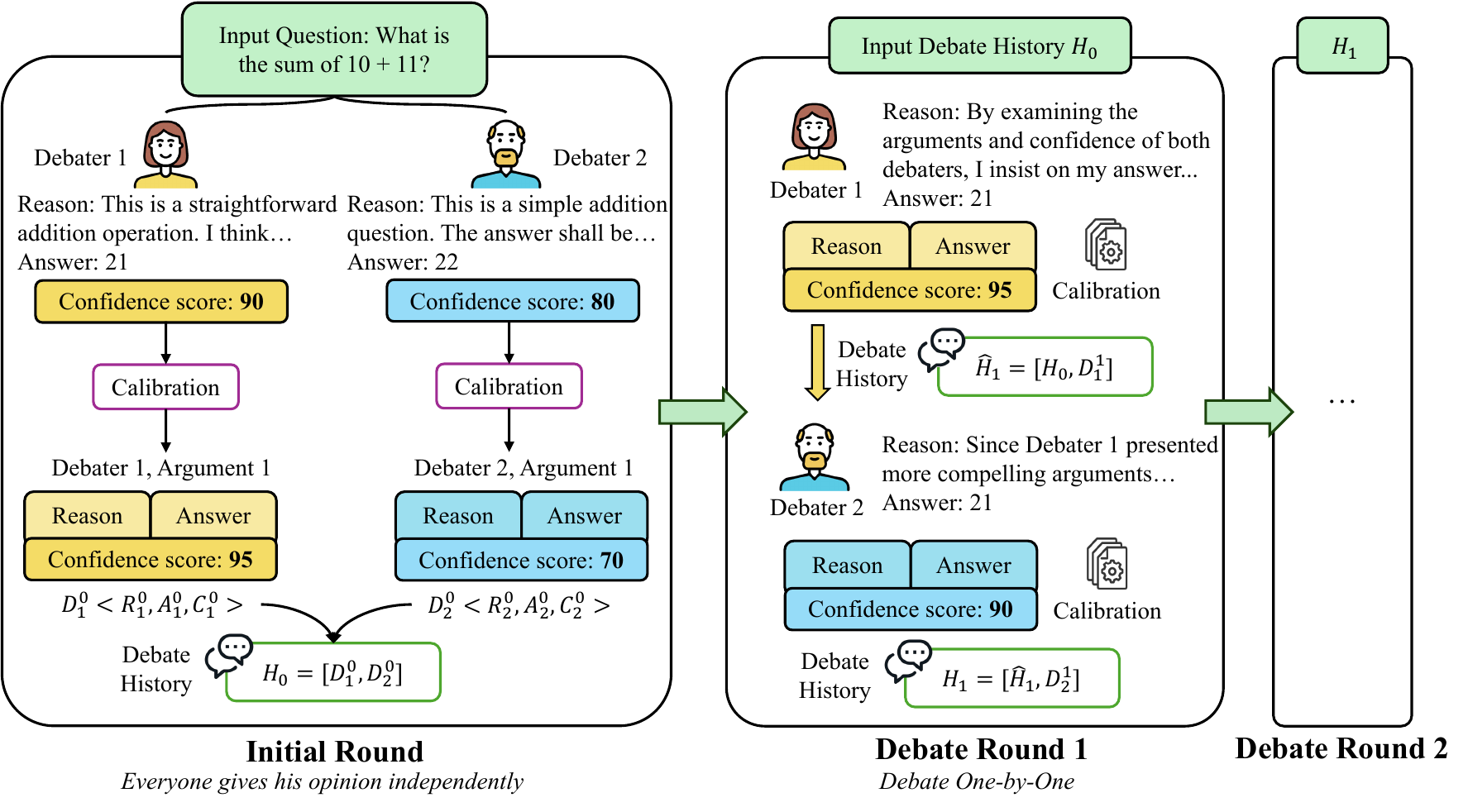}
  \caption{Overall design of ConfMAD. For a given question, debaters independently generate their Reason, Answer, and Confidence Score in the initial round ($r=0$), forming the initial debate history $H_0$. In subsequent rounds ($r > 0$), each debater reads the current history $H_r$, appends a new response, and updates the history. A round concludes once all debaters have responded.}
  \label{fig:ConfMAD}
\end{figure*}

In this section, we are going to introduce the design of ConfMAD. The overall design of ConfMAD is presented in Figure \ref{fig:ConfMAD}. The core design elements of ConfMAD encompass \textbf{Confidence Expression}, \textbf{Calibration Scheme}, and \textbf{Debate Workflow}.

\subsection{Confidence Expression}

There are multiple ways to elicit confidence from LLMs \citep{jiang2021can, si2022prompting, lin2022teaching, mielke2022reducing, xiong2023can, tian2023just, yang2024verbalized}. Referring to \citet{xiong2023can}, we use two simple and cost-effective methods to elicit each agent's confidence scores. The first method is Key Length-Normalized Sequence Probability Confidence (LN Confidence), and the second method is Self-Verbalized Confidence (SV Confidence):

\begin{itemize}[leftmargin=*]
    \item LN Confidence: Given a question, we compute the LN confidence score by first extracting the key tokens that form the final answer from the full output generated by the LLM. For instance, if the LLM outputs: "Reasoning: ... Answer: 14 7", we isolate "Answer: 14 7" as the answer tokens. We then calculate the probability of this token sequence and normalize it by its length, using the formula $seqprob^{1/n}$, where $n$ denotes the number of answer tokens.
    \item SV Confidence: Given a question, we prompt the LLM to generate not only its reasoning and answer but also a confidence score between 0 and 100. Specifically, the prompt encourages the model to append a confidence statement in the format: "Confidence: [Your confidence, 0–100]".
\end{itemize}

We also conducted a small-scale experiment to compare coarse and fine-grained confidence scores. Refer to Appendix~\ref{sec:categorical_confidence}.

\subsection{Calibration Scheme}

Since overconfidence is a prevalent issue in modern deep neural networks \citep{nguyen2015deep, guo2017calibration, kadavath2022language, mielke2022reducing, xiong2023can}, we introduce calibration methods. Specifically, we adopt three commonly used calibration schemes: Platt Scaling, Histogram Binning, and Temperature Scaling.

\paragraph{Platt Scaling} A simple and effective parametric calibration method that fits a logistic regression on validation data to map raw scores to calibrated probabilities using $P(y=1 | s) = \sigma(As + B)$, where $s$ is the raw score and $A$, $B$ are learned parameters \citep{platt1999probabilistic}.

\paragraph{Histogram Binning} A non-parametric method that partitions prediction scores into bins and assigns each bin a calibrated probability based on the empirical fraction of positive samples \cite{zadrozny2001obtaining}.

\paragraph{Temperature Scaling} A post-hoc calibration method that scales the logits by a temperature parameter $T > 0$ before applying softmax on the logits $\mathbf{z} = (z_1, z_2, ..., z_n)$ \citep{guo2017calibration}. The calibrated probabilities are computed as:
\begin{equation}
\label{eq:softmax_temp}
q_i = \frac{e^{z_i / T}}{\sum_{j=1}^{n} e^{z_j / T}}
\end{equation}
Temperature scaling is only applicable to LN confidence. We train the temperature parameter using the logits of the key tokens that  make up the answer.

Calibration models are trained before debates, so that we can load these models during debates to calibrate confidence scores. We randomly sample and separate an independent validation set from the dataset to train the calibration models. When training calibration models, we only conduct initial debate rounds, using the obtained answers and confidence scores.

\subsection{Debate Workflow}

Given a question $x$ with the correct answer $y$, our debate framework is designed with reference to the approach of \citet{du2023improving}, where agent communication occurs in a \textit{one-by-one} format. We adopt the one-by-one debate framework in order to enable a more direct comparison with the debate framework proposed by \citet{du2023improving}. For a MAD system consisting of $n$ LLM instances (Agents) from $M_1$ to $M_n$, when presented with question $x$:
\begin{itemize}[leftmargin=*]
    \item Round $r=0$ (Initial round): Each agent $M_i$ takes the input $x$ along with a prompt $p$ and produces a reason $R_i^0$, an answer $A_i^0$, and an original confidence score $C_{i}^{'0}$. We then apply a calibration model to transform $C_{i}^{'0}$ into a calibrated score $C_i^0$. This process is formally expressed as $M_i(x|p) = \langle R_i^0, A_i^0, C_i^0 \rangle = D_i^0$. The outputs from all agents $D_1^0, \dots, D_n^0$ are then concatenated to form the initial debate history $H_0$.
    \item Round $r>0$: Agent $M_1$ first receives the question $x$, prompt $p$, and debate history $H_{r-1}$, then provides new reasoning $R_1^r$, answer $A_1^r$ and confidence score $C_1^r$, expressed as $M_1(x|p,H_{r-1}) = \langle R_1^r, A_1^r, C_1^r \rangle = D_1^r$. Subsequently, we concatenate $D_1^r$ with $H_{r-1}$ to obtain the updated debate history $\hat{H}_r = \langle H_{r-1}, D_1^r \rangle$. The question, prompt, and debate history are then fed into $M_2$ for $D_2^r$. This process continues, appending new debate content to the history and querying each subsequent agent until all LLMs have participated, resulting in the complete debate history $H_r$.
\end{itemize}

The debate concludes after a predetermined number of rounds $T$. Figure \ref{fig:ConfMAD} illustrates this process of ConfMAD. Refer to Appendix~\ref{sec:design} for more details on the prompt design and the workflow of ConfMAD. Another debate mode, different from the one-by-one setting, is the \textit{broadcast} mode, where all debaters independently present their reasoning and answers in each debate round. Refer to Appendix~\ref{sec:onebyone_broadcast} for a detailed comparison.

\section{Experiments}

\subsection{Experiments Setup}
\paragraph{Benchmarks} We evaluated ConfMAD on four benchmarks, including BIGGSM \citep{chen2024unlocking}, Big-Bench-Hard \citep{suzgun2022challenging}, MMLU \citep{hendrycks2020mmlu}, and MATH \citep{hendrycks2021math}. BIGGSM is a collection of challenging mathematical computation problems that provide higher computational complexity and longer reasoning chains. To balance experimental reliability and overhead, we only sampled portions of MMLU, Big-Bench-Hard (BBH), and MATH to serve as our test and validation sets. The specific sizes are detailed in Appendix \ref{sec:size}, along with a discussion of the licensing terms for the datasets used.

\begin{table*}[!htbp]
  \centering
  \resizebox{\textwidth}{!}{
    \begin{tabular}{l|l|cc|cccc|ccc|cc}
      \toprule
      \multirow{1}{*}{\textbf{Partner}} & \multirow{2}{*}{\textbf{Task}} & \multicolumn{2}{c|}{\textbf{CoT}} & \multirow{2}{*}{\textbf{No Conf}} & \multirow{2}{*}{\textbf{Inter}} & \multirow{2}{*}{\textbf{CE}} & \multirow{2}{*}{\textbf{MP}} & \multicolumn{3}{c|}{\textbf{LN Confidence}} & \multicolumn{2}{c}{\textbf{SV Confidence}} \\
      4o-mini+ & & \textbf{4o-mini} & \textbf{Partner} & & & & & \textbf{Platt} & \textbf{Histo} & \textbf{Temp} & \textbf{Platt} & \textbf{Histo} \\
      \midrule
      \multirow{4}{*}{LLaMA} 
      & BIGGSM & 0.628 & 0.534 & 0.621 & 0.629 & 0.625 & 0.593 & \underline{0.632} & 0.625 & 0.627 & \textbf{0.655} & 0.620 \\
      & BBH    & 0.718 & 0.681 & 0.730 & 0.690 & 0.721 & 0.654 & \underline{0.763} & 0.753 & 0.751 & \textbf{0.767} & 0.759 \\
      & MMLU   & 0.763 & 0.820 & 0.783 & 0.736 & 0.780 & 0.747 & \textbf{0.833} & 0.829 & 0.824 & \underline{0.831} & 0.805 \\
      & MATH   & 0.600 & 0.534 & 0.693 & 0.665 & 0.691 & 0.717 & 0.711 & 0.710 & 0.711 & \textbf{0.725} & \underline{0.720} \\
      \midrule
      \multirow{4}{*}{Phi} 
      & BIGGSM & 0.628 & \textbf{0.760} & 0.693 & 0.670 & 0.630 & 0.590 & 0.747 & 0.735 & \underline{0.748} & 0.730 & 0.675 \\
      & BBH    & 0.718 & 0.709 & 0.738 & 0.711 & 0.693 & 0.669 & 0.777 & 0.753 & \underline{0.780} & \textbf{0.781} & 0.757 \\
      & MMLU   & 0.763 & 0.782 & 0.805 & 0.794 & 0.781 & 0.766 & \textbf{0.835} & \underline{0.834} & 0.833 & \underline{0.834} & 0.815 \\
      & MATH   & 0.600 & 0.730 & 0.755 & 0.765 & 0.693 & 0.744 & \textbf{0.785} & 0.765 & 0.782 & \underline{0.784} & 0.780 \\
      \bottomrule
    \end{tabular}
  }
  \caption{Comparative evaluation of accuracy between ConfMAD debate and baseline methods. The upper part presents experimental results derived from debates between 4o-mini and LLaMA, whereas the lower part presents the results obtained when 4o-mini engages with Phi. Bold and underlined values indicate the highest and second-highest accuracies in each row.}
  \label{tab:combined_debate_results}
\end{table*}

\paragraph{Models} We conducted debates on ConfMAD using two pairs of LLMs: GPT-4o-mini (referred to as 4o-mini) \citep{gpt4omini} with LLaMA-3.1-70B-Instruct (LLaMA) \citep{llama3.1}, and GPT-4o-mini with Phi-4 (Phi) \citep{phi4}.

\paragraph{Confidence Expression:} Confidence scores are expressed as scores ranging from 0 to 100. Consistent with our previous discussion, we employed two confidence expression methods: Key Length-Normalized Sequence Probability Confidence (LN) and Self-Verbalized Confidence (SV).

\paragraph{Calibration:} We included Vanilla confidence (i.e., without calibration) as one of our confidence expression variants. For LN confidence, we applied three calibration methods: Platt Scaling (Platt), Histogram Binning (Histo), and Temperature Scaling (Temp). For SV confidence, we used Platt Scaling and Histogram Binning.

\paragraph{Debate Setting} Our ConfMAD debate follows the one-by-one format proposed by \citet{du2023improving}, where agents communicate sequentially. Based on prior work \citep{liang2023encouraging, du2023improving, estornell2024multi} and our own experiments (see Appendix~\ref{sec:round}), debates typically converge within 2-3 rounds. Longer debates may introduce overly complex contexts, potentially harming performance. To balance effectiveness and cost, we adopt one initial round followed by two one-by-one rounds. In each debate round, 4o-mini speaks first, with LLaMA/Phi speaking afterwards. 

\paragraph{Baselines}

We selected five baselines for comparison with ConfMAD in our experiments:
\begin{itemize}[leftmargin=*]
  \item Chain-of-Thought (CoT): The method proposed by \citet{wei2022chain} encourages LLMs to output detailed reasoning steps when answering questions, which improves the performance of LLMs.
  \item No Confidence Debate (No Conf): This is similar to the MAD system proposed by \citet{du2023improving}. No Conf can be obtained by simply removing the confidence expression from ConfMAD.
  \item Interventions (Inter): \citet{estornell2024multi} introduced various intervention methods to improve the quality of debates in MAD systems, including diversity pruning, text quality pruning, and modification interventions.
  \item ChatEval (CE): \citet{chan2023chateval} investigated the impact of different communication settings and role assignments on MAD systems. In our experiments, we adopt the Simultaneous-Talk-with-Summarizer setting as a baseline method, where 4o-mini serves as the summarizer.
  \item Multi-Persona (MP): \citet{liang2023encouraging} introduced a MAD framework that incorporates an Affirmative Debater, a Negative Debater, and a Moderator to mitigate the issue of ‘thinking degradation.’ We assign 4o-mini as the Affirmative Debater, LLaMA/Phi-4 as the Negative Debater, and 4o-mini as the moderator.
\end{itemize} 

For No Conf and Inter, we use majority voting across agents, with ties resolved uniformly at random. When confidence scores are available, we select the answer from the highest-confidence agent (ties resolved uniformly at random). For CE and MP, we take the summarizer/moderator’s decision as the final answer.

\subsection{Results and Analysis}

In this section, we present our experimental results with ConfMAD on four benchmarks and discuss the following research questions: \textbf{RQ1:} How does ConfMAD perform compared to baseline methods?
\textbf{RQ2:} What is the impact of debating with confidence on individual LLMs' performance within MAD systems?
\textbf{RQ3:} How does debate with confidence improve MAD system performance?
\textbf{RQ4:} Ablation study on the impact of calibration schemes in ConfMAD.

\paragraph{RQ1: How does ConfMAD perform compared to baseline methods?} 
This question evaluates the overall effectiveness of our approach. Table~\ref{tab:combined_debate_results} presents the results across four selected benchmarks. In most cases, the highest and second-highest accuracies on each dataset are achieved under ConfMAD settings, highlighting the benefits of incorporating confidence expression into MAD systems. For the 4o-mini and LLaMA pairing, SV+Platt achieved the best accuracy on BIGGSM (0.655), outperforming No Conf and Inter by 5.5\% and 4.1\%, respectively. LN+Platt achieved the highest accuracy (0.833) on MMLU, exceeding No Conf by 6\%. For debates involving Phi, ConfMAD again yielded substantial gains. On BBH and MMLU, SV+Platt and LN+Temp ranked top, and both also led on MATH. While CoT alone produced the best score on BIGGSM, ConfMAD settings like LN+Platt (0.747) and LN+Temp (0.748) remained highly competitive, with higher accuracy than No Conf (0.693) and Inter (0.670). 

Among other baselines, the Inter setting often showed weaker performance, possibly due to its reliance on extensive pruning, which may be less effective in settings with a limited number of agents. CE consistently outperforms the No Conf setting on most datasets except BBH, but only matches the best ConfMAD variants (e.g., LN+Platt and LN+Temp) on MMLU. In contrast, MP performs significantly worse than all other methods, even underperforming the No Conf baseline. We observe that MP enforces the Negative side to explicitly oppose the Affirmative side’s answer, which leads the Negative agent to provide incorrect responses, even for questions it could originally answer correctly. This rigid assignment of affirmative and negative roles constrains the system’s ability to leverage the diverse knowledge and reasoning skills of heterogeneous agents. Furthermore, the agents’ stubbornness in defending incorrect answers harms the overall MAD performance. These findings further highlight the necessity of introducing calibrated confidence expression to improve coordination and decision-making in MAD systems. In summary, \emph{ConfMAD demonstrates superior performance compared to CoT in most scenarios and consistently outperforms baseline settings.}

\begin{table*}[!htbp]
  \centering
  \resizebox{0.85\textwidth}{!}{
  \begin{tabular}{l|l|ccc|ccc}
    \toprule
    \multirow{1}{*}{\textbf{Partner}} & \multirow{2}{*}{\textbf{Task}} & \multicolumn{3}{c|}{\textbf{4o-mini}} & \multicolumn{3}{c}{\textbf{Partner Model}} \\
    4o-mini+ & & \textbf{No Conf} & \textbf{LN+Platt} & \textbf{SV+Platt} & \textbf{No Conf} & \textbf{LN+Platt} & \textbf{SV+Platt} \\
    \midrule
    \multirow{4}{*}{LLaMA} 
    & BIGGSM & 0.619 & 0.628 & 0.644 & 0.625 & 0.633 & 0.635 \\
    & BBH    & 0.740 & 0.763 & 0.768 & 0.693 & 0.751 & 0.750 \\
    & MMLU   & 0.798 & 0.814 & 0.810 & 0.767 & 0.834 & 0.831 \\
    & MATH   & 0.706 & 0.720 & 0.725 & 0.680 & 0.704 & 0.732 \\
    \midrule
    \multirow{4}{*}{Phi} 
    & BIGGSM & 0.655 & 0.733 & 0.718 & 0.690 & 0.755 & 0.738 \\
    & BBH    & 0.757 & 0.777 & 0.778 & 0.719 & 0.762 & 0.755 \\
    & MMLU   & 0.806 & 0.833 & 0.830 & 0.804 & 0.827 & 0.838 \\
    & MATH   & 0.748 & 0.774 & 0.766 & 0.771 & 0.784 & 0.784 \\
    \bottomrule
  \end{tabular}
  }
  \caption{Final-round accuracy comparison of debaters under different model pairings and debate settings. Only results for Platt setting are reported in this table. The upper section presents results for debates between 4o-mini and LLaMA, while the lower section presents results for 4o-mini and Phi.}
  \label{tab:FinalAcc}
\end{table*}

\paragraph{RQ2: What is the impact of debating with confidence on individual LLM's performance within MAD systems?} 

This research question primarily focuses on the impact of introducing confidence expression on the final performance of each individual LLM agent participating in debates. We compared the final accuracy of each LLM with its accuracy under the No Conf setting. The results are presented in Table~\ref{tab:FinalAcc}.

We observe that after introducing confidence scores, individual LLM performance shows significant improvement. For instance, when 4o-mini and LLaMA debate on BBH, applying LN+Platt and SV+Platt enables LLaMA to achieve final-round accuracies of 0.751 and 0.750 respectively, which substantially exceed the 0.693 accuracy observed under the No Conf setting. These results suggest that \emph{ConfMAD does not simply select the answer of the stronger LLM based on confidence scores.} Instead, it leverages these scores to guide participating LLMs toward deeper reasoning and self-evaluation. This facilitates collaborative interaction in which all debaters benefit, ultimately improving the overall performance of MAD systems.

\paragraph{RQ3: How does debate with confidence improve MAD system performance?} 

\begin{figure}[ht]
  \includegraphics[width=\columnwidth]{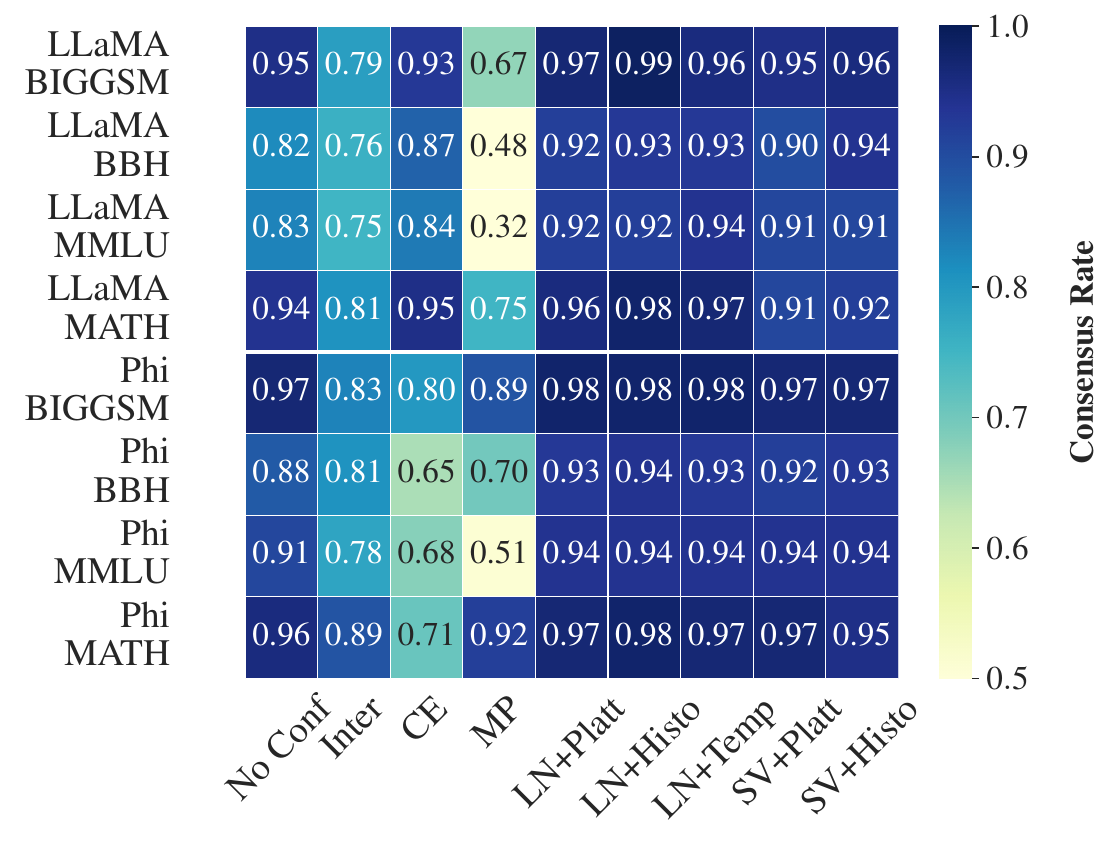}
  \caption{The ratio of cases reaching consensus after three rounds of debate under different debate settings across selected benchmarks.}
  \label{fig:Heatmap}
\end{figure}

\begin{figure*}[t]
  \centering
  \includegraphics[width=\textwidth]{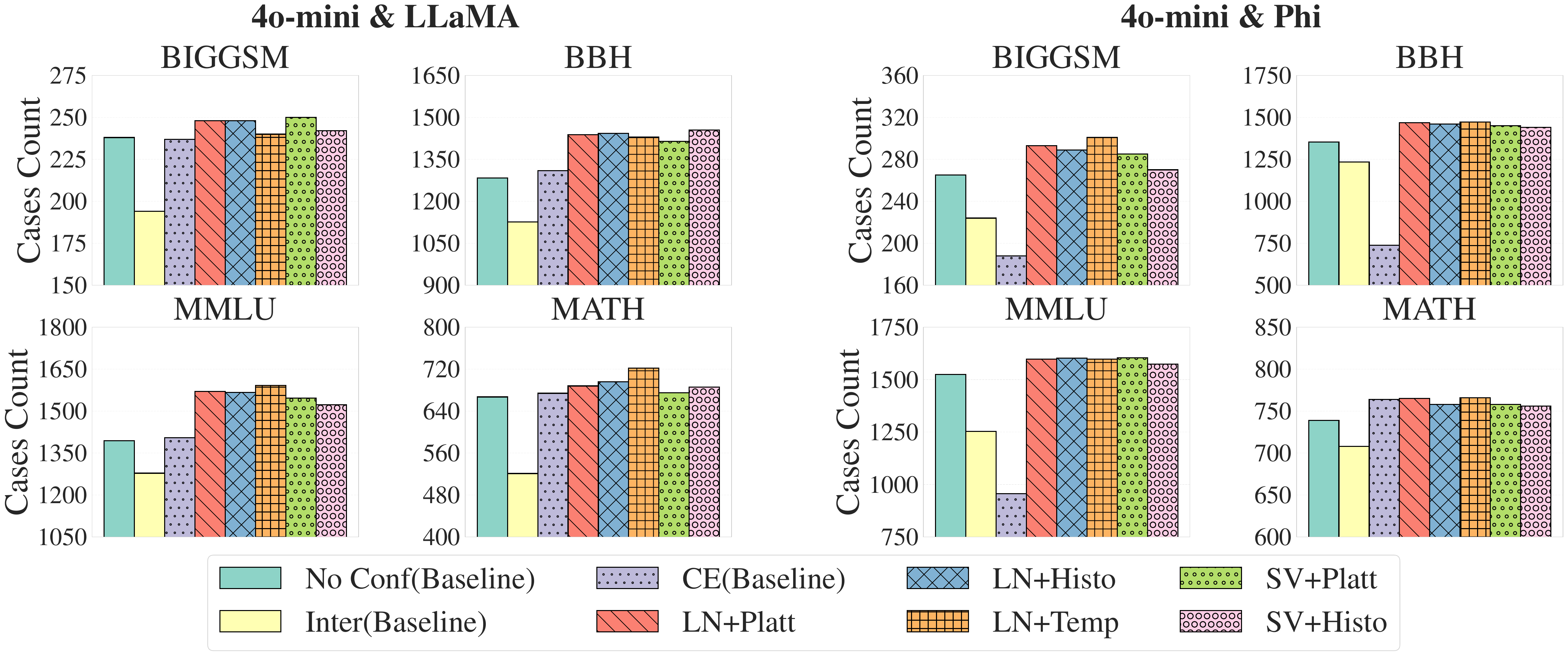}
  \caption{The number of cases achieving correct consensus across different debate settings and datasets. Figures on the left show debates between 4o-mini and LLaMA, while the right ones present debates between 4o-mini and Phi. We do not report MP results here, as its side configuration yielded too few consensus cases on some tasks.}
  \label{fig:AllRight}
\end{figure*}

In this research question, we analyze specifically how confidence debate enhances the performance of MAD systems. \emph{We find that introducing confidence scores significantly helps MAD systems reach consensus and reach correct agreements.} Figure \ref{fig:Heatmap} illustrates the ratio of cases where consensus was reached after debate, comparing baseline debate methods with various ConfMAD settings. The results show that across almost all settings, debates using ConfMAD achieve higher consensus rates, with particularly notable improvements on the BBH and MMLU datasets. For instance, when 4o-mini debates with LLaMA on the MMLU dataset, the consensus rate increases by around 11\%. Figure \ref{fig:AllRight} further demonstrates the number of cases where correct consensus was achieved (where both debating agents ultimately provide the correct answer) under baseline methods versus different ConfMAD settings. In most scenarios, ConfMAD yields more correct consensus cases than the baseline methods. 

We also observed a phenomenon we refer to as \textit{Correction}, where the debate process leads to a correct final answer even though at least one agent initially provides an incorrect response. We found that debates with confidence substantially increase the number of correction cases across most settings. For example, when 4o-mini debates with Phi on the MMLU dataset, the number of correction cases increases by approximately 20\%. Refer to Appendix \ref{sec:correction} for more details.

\begin{table*}[ht]
  \centering
  \resizebox{0.8\textwidth}{!}{
  \begin{tabular}{l|l|cccc|ccc}
    \toprule
    \multirow{1}{*}{\textbf{Partner}} & \multirow{2}{*}{\textbf{Task}} & \multicolumn{4}{c|}{\textbf{LN}} & \multicolumn{3}{c}{\textbf{SV}} \\
    4o-mini+ & & \textbf{Vanilla} & \textbf{Platt} & \textbf{Histo} & \textbf{Temp} & \textbf{Vanilla} & \textbf{Platt} & \textbf{Histo} \\
    \midrule
    \multirow{4}{*}{LLaMA} 
    & BIGGSM & 0.627 & \underline{0.632} & 0.625 & 0.627 & 0.608 & \textbf{0.655} & 0.620 \\
    & BBH    & 0.748 & \underline{0.763} & 0.753 & 0.751 & \underline{0.763} & \textbf{0.767} & 0.759 \\
    & MMLU   & 0.804 & \textbf{0.833} & 0.829 & 0.824 & 0.829 & \underline{0.831} & 0.805 \\
    & MATH   & 0.719 & 0.711 & 0.710 & 0.711 & \textbf{0.730} & \underline{0.725} & 0.720 \\
    \midrule
    \multirow{4}{*}{Phi} 
    & BIGGSM & 0.694 & \underline{0.747} & 0.735 & \textbf{0.748} & 0.708 & 0.730 & 0.675 \\
    & BBH    & 0.778 & 0.777 & 0.753 & \underline{0.780} & 0.760 & \textbf{0.781} & 0.757 \\
    & MMLU   & 0.801 & \textbf{0.835} & \underline{0.834} & 0.833 & 0.829 & \underline{0.834} & 0.815 \\
    & MATH   & \underline{0.784} & \textbf{0.785} & 0.765 & 0.782 & \underline{0.784} & \underline{0.784} & 0.780 \\
    \bottomrule
  \end{tabular}
  }
  \caption{Comparison of accuracy between calibrated and uncalibrated (Vanilla) confidences across different debate settings. Bold and underlined values indicate the highest and second-highest accuracies in each row.}
  \label{tab:Ablation}
\end{table*}

\begin{figure}[t]
  \includegraphics[width=\columnwidth]{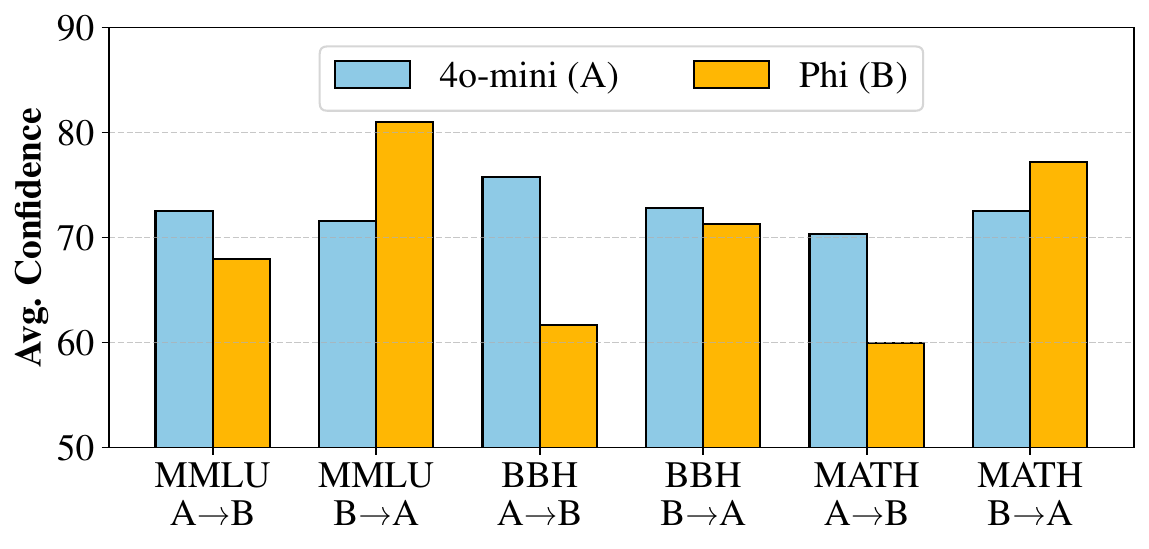}
  \caption{Average initial-round confidence scores when correction occurs. Arrows indicate correction direction (e.g., 4o-mini$\rightarrow$Phi means 4o-mini corrected Phi).}
  \label{fig:ConfCompare}
\end{figure}

We further analyzed LLMs' confidence in the initial round for cases involving correction. Figure~\ref{fig:ConfCompare} shows the average confidence of 4o-mini and Phi when correcting each other. On MMLU, when 4o-mini corrected Phi, its average confidence was 72.5 compared to Phi's 68.0. In contrast, when Phi corrected 4o-mini, its confidence was 81.0 while 4o-mini's was only 71.6. On BBH, 4o-mini had higher average confidence in both directions, but the gap narrowed when Phi corrected 4o-mini. These patterns highlight the role of confidence scores in guiding the MAD system toward correct consensus.

\paragraph{RQ4: Ablation study on the impact of calibration schemes in ConfMAD.}

In this section, we conduct an ablation study to examine the effect of calibration schemes on the performance of ConfMAD. We disable calibration and rerun experiments on the same datasets; the results are presented in Table~\ref{tab:Ablation}. We find that most of the highest and second-highest accuracies are achieved using Platt Scaling, followed by Temperature Scaling. The results also indicate that disabling calibration does not consistently degrade performance; in some cases, the accuracy remains comparable or even slightly improves. Additionally, we observe that Histogram Binning exhibits highly unstable performance and frequently underperforms. \emph{These findings suggest that Platt Scaling is a relatively robust calibration method, while the effectiveness of Histogram Binning remains questionable.}

\begin{table}[!htbp]
  \centering
  \resizebox{\columnwidth}{!}{
    \begin{tabular}{lccc}
      \toprule
      \textbf{Setting} & \textbf{Round 0} & \textbf{Round 1}  & \textbf{Accuracy} \\
      \midrule
      \multicolumn{4}{c}{MMLU, 4o-mini \& Phi} \\
      \midrule
      LN+Platt   & 0.668 (263/394) & 0.575 (100/174) & 0.835 \\
      % LN+Histo   & 0.683 (231/338) & 0.585 (86/147)  & 0.834 \\
      LN+Temp    & 0.683 (231/338) & 0.585 (86/147)  & 0.833 \\
      LN+Vanilla & 0.608 (129/212) & 0.273 (27/99)   & 0.801 \\
      \midrule
      \multicolumn{4}{c}{BBH, 4o-mini \& Phi} \\
      \midrule
      LN+Platt & 0.614 (316/515) & 0.514 (111/216) & 0.777 \\
      LN+Histo & 0.552 (280/507) & 0.488 (81/166)  & 0.753 \\
      LN+Temp  & 0.658 (324/500) & 0.550 (120/218) & 0.780 \\
      \bottomrule
    \end{tabular}
  }
  \caption{Win Rate (WR) across debate settings on MMLU and BBH. WR denotes the proportion of cases where the correct LLM had a higher confidence score than the incorrect one, in instances where both the answers and confidence scores differ.}
  \label{tab:WinRate}
\end{table}

\begin{figure}[t]
  \includegraphics[width=\columnwidth]{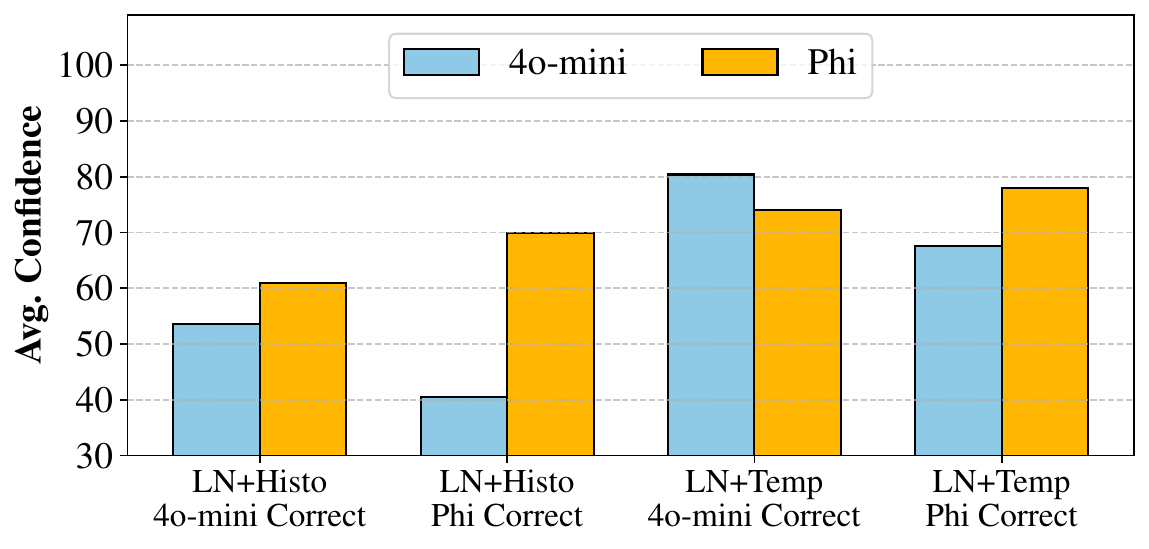}
  \caption{Average confidence scores in the initial round when only one LLM (4o-mini or Phi) gives the correct answer. For example, "4o-mini correct" indicates that 4o-mini is correct while Phi is incorrect.}
  \label{fig:WRConf}
\end{figure}

To further explore how confidence scores affect MAD systems, we analyze some representative cases and offer corresponding insights. A notable example is the debate between 4o-mini and Phi on MMLU, where LN+Platt and LN+Vanilla settings achieved final accuracies of 0.835 and 0.801, respectively, despite using the same LN confidence mode. We then analyzed the ratio of cases where the confidence score of the correct LLM agent prevailed when answers disagreed (as Win Rate, WR). As shown in Table \ref{tab:WinRate}, in the LN+Vanilla setting, raw confidence scores often failed to accurately reflect one agent's knowledge advantage over another, affecting both the debate process and the final confidence-based answer selection. Table~\ref{tab:WinRate} also presents debate results on BBH using 4o-mini and Phi under the LN confidence mode. While most settings achieved final accuracies around 0.780, LN+Histo lagged behind at 0.753 and had the lowest first-round WR at 0.552. We find that this may be due to Histogram Binning excessively downscaling the confidence scores. Figure~\ref{fig:WRConf} shows that under LN+Histo, even when 4o-mini was correct, its confidence was substantially lower than Phi’s, indicating a misalignment between correctness and confidence. 

Based on the previous observations, we hypothesize that the performance of MAD systems may benefit when confidence scores effectively reflect the relative capabilities of LLM agents. Specifically, we believe two factors are essential for this to hold. First, confidence scores should accurately capture each agent’s underlying knowledge level with respect to a given query. Second, appropriate calibration methods are needed to align the confidence scores of different LLMs onto a comparable scale, enabling meaningful and reliable comparison across agents.

% Histogram Binning in this case overly downscaled 4o-mini’s confidence, whereas Platt and Temperature Scaling produced more reliable estimates. Although LN+Histo led to more correct consensus cases—likely due to Phi's high confidence—it also increased incorrect consensus, possibly because 4o-mini remained underconfident even when correct. These results suggest that when calibrated confidence fails to align with correctness, debate performance may decline. Since calibration models are trained on raw confidence scores, their effectiveness depends on both the calibration method and the quality of the underlying confidence elicitation.

% From the above discussion, we derive an instructive insight: well-expressed confidence can further enhance the performance of MAD systems. This involves two key aspects:
% \begin{itemize}[leftmargin=*]
%     \item \textbf{Confidence quality of LLMs:} For any given query, the confidence expressed by an LLM should faithfully reflect its actual knowledge about the question. That is, when the LLM possesses relevant knowledge, it should exhibit high confidence, and when its knowledge is lacking, it should express low confidence accordingly.
    
%     \item \textbf{Comparability of confidence across multiple LLMs:} Different LLMs may have varying tendencies in expressing confidence (overconfident or underconfident). Therefore, appropriate calibration methods are needed to align their confidence scores on a common scale, enabling meaningful comparisons across different LLM agents.
% \end{itemize}

We present more details and discussions about our experimental results in Appendix \ref{sec:extended}.

\section{Related Works}

\paragraph{Multi-Agent Debate Systems}

Multi-Agent Debate (MAD) systems improve response quality by simulating debates among LLMs. \citet{du2023improving} first introduced this framework and demonstrated its effectiveness across multiple benchmarks. Subsequent work expanded it in various directions: \citet{liang2023encouraging}, \citet{chan2023chateval}, and \citet{li2023camel} assigned diverse roles (e.g., judges, professionals) to agents; \citet{khan2024debating} showed that more persuasive debaters yield more accurate responses; \citet{wang2023can} assessed models' ability to defend truth, while \citet{taubenfeld2024systematic} identified systemic biases such as position and first-mover effects. \citet{estornell2024multi} proposed interventions like diversity pruning and misunderstanding refutation, and \citet{li2024improving}, \citet{liu2024groupdebate} optimized debate topologies to reduce computation. Our work contributes to this line by enhancing MAD performance through the simple incorporation of confidence scores.

\paragraph{Confidence Expression}

Several studies have explored how to elicit confidence from LLMs. \citet{jiang2021can}, \citet{si2022prompting}, and \citet{xiong2023can} proposed using token-normalized probabilities to quantify confidence and uncertainty. \citet{lin2022teaching} introduced verbalized confidence through fine-tuning, prompting models to explicitly express confidence levels. This direction was further extended by \citet{xiong2023can}, \citet{yang2024verbalized}, and \citet{tian2023just} under various prompting strategies. Other uncertainty estimation methods have also been applied to confidence expression, such as semantic entropy from \citet{kuhn2023semantic}, which uses sampling-based estimation, and internal representation-based approaches from \citet{mielke2022reducing} and \citet{gao2025flue}.

\paragraph{Calibration} 

Modern deep neural networks often suffer from poor calibration (\citet{guo2017calibration}, \citet{minderer2021revisiting}, \citet{xiong2023proximity}). Researchers have developed various post-processing calibration techniques. These methods primarily fall into two categories: parametric scaling methods (\citet{platt1999probabilistic}, \citet{guo2017calibration}, and \citet{deng2023great}) and non-parametric binning methods (\citet{zadrozny2001obtaining}, \citet{zhang2020mix}). In this study, we adopt three representative calibration methods: Platt Scaling \citep{platt1999probabilistic}, non-parametric Histogram Binning \citep{zadrozny2001obtaining}, and Temperature Scaling \citep{guo2017calibration}, which are tailored for deep neural networks.

Some prior work has also explored the use of uncertainty metrics or confidence scores in MAD systems, but without a systematic study of how explicit confidence can be expressed and calibrated. ConfidenceCal \citep{2024ConfidenceCal} focuses on reweighting tokens from different debaters using uncertainty estimates. RECONCILE \citep{2023ReConcile} encourages LLMs to output confidence via prompting and applies a fixed post-hoc calibration, while overlooking potential differences in knowledge scope and reliability across models. In this paper, we provide the first systematic comparison of multiple forms of confidence expression and different calibration methods. We analyze their influence on MAD dynamics in detail. By incorporating calibrated confidence expressions, we improve the accuracy of MAD systems as well as their robustness.

\section{Conclusion}
In this work, we propose incorporating confidence expression into MAD systems to enhance their performance. We explore two confidence expression methods and several calibration schemes to adjust confidence scores. To validate our approach, we develop \textbf{ConfMAD}, a confidence-aware MAD framework. Experiments across multiple benchmarks show that confidence expression not only improves the accuracy of individual LLM agents but also helps the system reach correct consensus more effectively. We further analyze the impact of different calibration methods and highlight the importance of aligning confidence scores across agents. Our findings underscore the critical role of well-calibrated confidence in enabling more reliable and effective multi-agent debate and offer insights for designing better confidence mechanisms in future MAD systems.

\section*{Limitations}

In this paper, we proposed using confidence expression in MAD systems and developed ConfMAD. We demonstrated the effectiveness of our approach through evaluations across various benchmarks. However, there are still some limitations to consider.

First, the confidence elicitation methods we currently adopted are relatively simple and may suffer from issues such as instability across repeated queries and overly concentrated distributions. While more advanced elicitation techniques exist, many rely on multiple sampling, and MAD debates already involve repeated inference of LLMs. This makes efficiency optimization an important direction for future work. In addition, more advanced applications of confidence scoring, such as step-wise expression in multi-step reasoning, have not yet been explored in this work.

Second, the generalization capacity of incorporating confidence expression in MAD systems, and more broadly in multi-agent settings, requires further investigation. Our current calibration design is still dataset-agnostic, which may limit generalization and necessitate retraining on datasets with different distributions. Moreover, the potential generalization ability of a calibration model trained on a specific dataset remains unexplored. We may also consider incorporating confidence calibration into more multi-agent systems to explore the generalization ability of the method we proposed.

Finally, certain design choices of ConfMAD could be further extended. For example, a dynamic stopping criterion could be employed instead of a fixed number of debate rounds. Alternative communication strategies, such as broadcasting modes rather than sequential one-by-one debates, also represent promising avenues for future research.

\section*{Ethical Considerations}

This work aims to improve the effectiveness and reliability of multi-agent debate (MAD) systems by incorporating confidence expression. All experiments are conducted on publicly available benchmark datasets without involving any personally identifiable information or human subjects. While our system encourages more truthful consensus through structured agent interaction, we acknowledge the potential risks of reinforcing model biases or producing persuasive yet incorrect outputs. To mitigate these concerns, we emphasize transparency in model design and calibration, and advocate for responsible deployment, particularly in high-stakes or decision-critical contexts. We reserve the right to restrict the use of this technology in applications that may lead to unethical outcomes or misuse.

\bibliography{main}

\begin{thebibliography}{41}
\providecommand{\natexlab}[1]{#1}

\bibitem[{Bai(2024)}]{2024ConfidenceCal}
Yilin Bai. 2024.
\newblock Confidencecal: Enhancing llms reliability through confidence calibration in multi-agent debate.
\newblock In \emph{2024 10th International Conference on Big Data and Information Analytics (BigDIA)}, pages 221--226.

\bibitem[{Chan et~al.(2023)Chan, Chen, Su, Yu, Xue, Zhang, Fu, and Liu}]{chan2023chateval}
Chi-Min Chan, Weize Chen, Yusheng Su, Jianxuan Yu, Wei Xue, Shanghang Zhang, Jie Fu, and Zhiyuan Liu. 2023.
\newblock Chateval: Towards better llm-based evaluators through multi-agent debate.
\newblock \emph{arXiv preprint arXiv:2308.07201}.

\bibitem[{Chen et~al.(2023)Chen, Saha, and Bansal}]{2023ReConcile}
Chih~Yao Chen, Swarnadeep Saha, and Mohit Bansal. 2023.
\newblock Reconcile: Round-table conference improves reasoning via consensus among diverse llms.

\bibitem[{Chen et~al.(2024)Chen, Qin, Wang, Zhou, and Che}]{chen2024unlocking}
Qiguang Chen, Libo Qin, Jiaqi Wang, Jingxuan Zhou, and Wanxiang Che. 2024.
\newblock Unlocking the capabilities of thought: A reasoning boundary framework to quantify and optimize chain-of-thought.
\newblock \emph{Advances in Neural Information Processing Systems}, 37:54872--54904.

\bibitem[{Deng et~al.(2023)Deng, Xiong, and Hooi}]{deng2023great}
Ailin Deng, Miao Xiong, and Bryan Hooi. 2023.
\newblock Great models think alike: improving model reliability via inter-model latent agreement.
\newblock \emph{arXiv preprint arXiv:2305.01481}.

\bibitem[{Du et~al.(2023)Du, Li, Torralba, Tenenbaum, and Mordatch}]{du2023improving}
Yilun Du, Shuang Li, Antonio Torralba, Joshua~B Tenenbaum, and Igor Mordatch. 2023.
\newblock Improving factuality and reasoning in language models through multiagent debate.
\newblock In \emph{Forty-first International Conference on Machine Learning}.

\bibitem[{Estornell and Liu(2024)}]{estornell2024multi}
Andrew Estornell and Yang Liu. 2024.
\newblock Multi-llm debate: Framework, principals, and interventions.
\newblock \emph{Advances in Neural Information Processing Systems}, 37:28938--28964.

\bibitem[{Gao et~al.(2025)Gao, Gong, Lin, Xu, Zhou, and Li}]{gao2025flue}
Shiqi Gao, Tianxiang Gong, Zijie Lin, Runhua Xu, Haoyi Zhou, and Jianxin Li. 2025.
\newblock Flue: Streamlined uncertainty estimation for large language models.
\newblock In \emph{Proceedings of the AAAI Conference on Artificial Intelligence}, volume~39, pages 16745--16753.

\bibitem[{Guo et~al.(2017)Guo, Pleiss, Sun, and Weinberger}]{guo2017calibration}
Chuan Guo, Geoff Pleiss, Yu~Sun, and Kilian~Q Weinberger. 2017.
\newblock On calibration of modern neural networks.
\newblock In \emph{International conference on machine learning}, pages 1321--1330. PMLR.

\bibitem[{Hendrycks et~al.(2020)Hendrycks, Burns, Basart, Zou, Mazeika, Song, and Steinhardt}]{hendrycks2020mmlu}
Dan Hendrycks, Collin Burns, Steven Basart, Andy Zou, Mantas Mazeika, Dawn Song, and Jacob Steinhardt. 2020.
\newblock Measuring massive multitask language understanding.
\newblock \emph{arXiv preprint arXiv:2009.03300}.

\bibitem[{Hendrycks et~al.(2021)Hendrycks, Burns, Kadavath, Arora, Basart, Tang, Song, and Steinhardt}]{hendrycks2021math}
Dan Hendrycks, Collin Burns, Saurav Kadavath, Akul Arora, Steven Basart, Eric Tang, Dawn Song, and Jacob Steinhardt. 2021.
\newblock Measuring mathematical problem solving with the math dataset.
\newblock \emph{arXiv preprint arXiv:2103.03874}.

\bibitem[{Jiang et~al.(2021)Jiang, Araki, Ding, and Neubig}]{jiang2021can}
Zhengbao Jiang, Jun Araki, Haibo Ding, and Graham Neubig. 2021.
\newblock How can we know when language models know? on the calibration of language models for question answering.
\newblock \emph{Transactions of the Association for Computational Linguistics}, 9:962--977.

\bibitem[{Kadavath et~al.(2022)Kadavath, Conerly, Askell, Henighan, Drain, Perez, Schiefer, Hatfield-Dodds, DasSarma, Tran-Johnson et~al.}]{kadavath2022language}
Saurav Kadavath, Tom Conerly, Amanda Askell, Tom Henighan, Dawn Drain, Ethan Perez, Nicholas Schiefer, Zac Hatfield-Dodds, Nova DasSarma, Eli Tran-Johnson, and 1 others. 2022.
\newblock Language models (mostly) know what they know.
\newblock \emph{arXiv preprint arXiv:2207.05221}.

\bibitem[{Khan et~al.(2024)Khan, Hughes, Valentine, Ruis, Sachan, Radhakrishnan, Grefenstette, Bowman, Rockt{\"a}schel, and Perez}]{khan2024debating}
Akbir Khan, John Hughes, Dan Valentine, Laura Ruis, Kshitij Sachan, Ansh Radhakrishnan, Edward Grefenstette, Samuel~R Bowman, Tim Rockt{\"a}schel, and Ethan Perez. 2024.
\newblock Debating with more persuasive llms leads to more truthful answers.
\newblock \emph{arXiv preprint arXiv:2402.06782}.

\bibitem[{Kuhn et~al.(2023)Kuhn, Gal, and Farquhar}]{kuhn2023semantic}
Lorenz Kuhn, Yarin Gal, and Sebastian Farquhar. 2023.
\newblock Semantic uncertainty: Linguistic invariances for uncertainty estimation in natural language generation.
\newblock \emph{arXiv preprint arXiv:2302.09664}.

\bibitem[{Li et~al.(2023)Li, Hammoud, Itani, Khizbullin, and Ghanem}]{li2023camel}
Guohao Li, Hasan Hammoud, Hani Itani, Dmitrii Khizbullin, and Bernard Ghanem. 2023.
\newblock Camel: Communicative agents for" mind" exploration of large language model society.
\newblock \emph{Advances in Neural Information Processing Systems}, 36:51991--52008.

\bibitem[{Li et~al.(2024)Li, Du, Zhang, Hou, Grabowski, Li, and Ie}]{li2024improving}
Yunxuan Li, Yibing Du, Jiageng Zhang, Le~Hou, Peter Grabowski, Yeqing Li, and Eugene Ie. 2024.
\newblock Improving multi-agent debate with sparse communication topology.
\newblock \emph{arXiv preprint arXiv:2406.11776}.

\bibitem[{Liang et~al.(2023)Liang, He, Jiao, Wang, Wang, Wang, Yang, Shi, and Tu}]{liang2023encouraging}
Tian Liang, Zhiwei He, Wenxiang Jiao, Xing Wang, Yan Wang, Rui Wang, Yujiu Yang, Shuming Shi, and Zhaopeng Tu. 2023.
\newblock Encouraging divergent thinking in large language models through multi-agent debate.
\newblock \emph{arXiv preprint arXiv:2305.19118}.

\bibitem[{Lin et~al.(2022)Lin, Hilton, and Evans}]{lin2022teaching}
Stephanie Lin, Jacob Hilton, and Owain Evans. 2022.
\newblock Teaching models to express their uncertainty in words.
\newblock \emph{arXiv preprint arXiv:2205.14334}.

\bibitem[{Liu et~al.(2024)Liu, Wang, Huang, Xu, Zeng, Jiang, Yang, and Li}]{liu2024groupdebate}
Tongxuan Liu, Xingyu Wang, Weizhe Huang, Wenjiang Xu, Yuting Zeng, Lei Jiang, Hailong Yang, and Jing Li. 2024.
\newblock Groupdebate: Enhancing the efficiency of multi-agent debate using group discussion.
\newblock \emph{arXiv preprint arXiv:2409.14051}.

\bibitem[{Madaan et~al.(2023)Madaan, Tandon, Gupta, Hallinan, Gao, Wiegreffe, Alon, Dziri, Prabhumoye, Yang et~al.}]{madaan2023self}
Aman Madaan, Niket Tandon, Prakhar Gupta, Skyler Hallinan, Luyu Gao, Sarah Wiegreffe, Uri Alon, Nouha Dziri, Shrimai Prabhumoye, Yiming Yang, and 1 others. 2023.
\newblock Self-refine: Iterative refinement with self-feedback.
\newblock \emph{Advances in Neural Information Processing Systems}, 36:46534--46594.

\bibitem[{{Meta}(2024)}]{llama3.1}
{Meta}. 2024.
\newblock \href {https://ai.meta.com/blog/meta-llama-3-1/} {Introducing llama 3.1: Our most capable models to date}.

\bibitem[{{Microsoft}(2025)}]{phi4}
{Microsoft}. 2025.
\newblock \href {https://techcommunity.microsoft.com/blog/aiplatformblog/introducing-phi-4-microsoft%E2%80%99s-newest-small-language-model-specializing-in-comple/4357090} {Introducing phi-4: Microsoft’s newest small language model specializing in complex reasoning}.

\bibitem[{Mielke et~al.(2022)Mielke, Szlam, Dinan, and Boureau}]{mielke2022reducing}
Sabrina~J Mielke, Arthur Szlam, Emily Dinan, and Y-Lan Boureau. 2022.
\newblock Reducing conversational agents’ overconfidence through linguistic calibration.
\newblock \emph{Transactions of the Association for Computational Linguistics}, 10:857--872.

\bibitem[{Minderer et~al.(2021)Minderer, Djolonga, Romijnders, Hubis, Zhai, Houlsby, Tran, and Lucic}]{minderer2021revisiting}
Matthias Minderer, Josip Djolonga, Rob Romijnders, Frances Hubis, Xiaohua Zhai, Neil Houlsby, Dustin Tran, and Mario Lucic. 2021.
\newblock Revisiting the calibration of modern neural networks.
\newblock \emph{Advances in neural information processing systems}, 34:15682--15694.

\bibitem[{Minsky(1986)}]{minsky1986society}
Marvin Minsky. 1986.
\newblock \emph{Society of mind}.
\newblock Simon and Schuster.

\bibitem[{Nguyen et~al.(2015)Nguyen, Yosinski, and Clune}]{nguyen2015deep}
Anh Nguyen, Jason Yosinski, and Jeff Clune. 2015.
\newblock Deep neural networks are easily fooled: High confidence predictions for unrecognizable images.
\newblock In \emph{Proceedings of the IEEE conference on computer vision and pattern recognition}, pages 427--436.

\bibitem[{{OpenAI}(2024)}]{gpt4omini}
{OpenAI}. 2024.
\newblock \href {https://openai.com/index/gpt-4o-mini-advancing-cost-efficient-intelligence/} {Gpt-4o mini: advancing cost-efficient intelligence}.

\bibitem[{Platt et~al.(1999)}]{platt1999probabilistic}
John Platt and 1 others. 1999.
\newblock Probabilistic outputs for support vector machines and comparisons to regularized likelihood methods.
\newblock \emph{Advances in large margin classifiers}, 10(3):61--74.

\bibitem[{Si et~al.(2022)Si, Gan, Yang, Wang, Wang, Boyd-Graber, and Wang}]{si2022prompting}
Chenglei Si, Zhe Gan, Zhengyuan Yang, Shuohang Wang, Jianfeng Wang, Jordan Boyd-Graber, and Lijuan Wang. 2022.
\newblock Prompting gpt-3 to be reliable.
\newblock \emph{arXiv preprint arXiv:2210.09150}.

\bibitem[{Suzgun et~al.(2022)Suzgun, Scales, Sch{\"a}rli, Gehrmann, Tay, Chung, Chowdhery, Le, Chi, Zhou et~al.}]{suzgun2022challenging}
Mirac Suzgun, Nathan Scales, Nathanael Sch{\"a}rli, Sebastian Gehrmann, Yi~Tay, Hyung~Won Chung, Aakanksha Chowdhery, Quoc~V Le, Ed~H Chi, Denny Zhou, and 1 others. 2022.
\newblock Challenging big-bench tasks and whether chain-of-thought can solve them.
\newblock \emph{arXiv preprint arXiv:2210.09261}.

\bibitem[{Taubenfeld et~al.(2024)Taubenfeld, Dover, Reichart, and Goldstein}]{taubenfeld2024systematic}
Amir Taubenfeld, Yaniv Dover, Roi Reichart, and Ariel Goldstein. 2024.
\newblock Systematic biases in llm simulations of debates.
\newblock \emph{arXiv preprint arXiv:2402.04049}.

\bibitem[{Tian et~al.(2023)Tian, Mitchell, Zhou, Sharma, Rafailov, Yao, Finn, and Manning}]{tian2023just}
Katherine Tian, Eric Mitchell, Allan Zhou, Archit Sharma, Rafael Rafailov, Huaxiu Yao, Chelsea Finn, and Christopher~D Manning. 2023.
\newblock Just ask for calibration: Strategies for eliciting calibrated confidence scores from language models fine-tuned with human feedback.
\newblock \emph{arXiv preprint arXiv:2305.14975}.

\bibitem[{Wang et~al.(2023)Wang, Yue, and Sun}]{wang2023can}
Boshi Wang, Xiang Yue, and Huan Sun. 2023.
\newblock Can chatgpt defend its belief in truth? evaluating llm reasoning via debate.
\newblock \emph{arXiv preprint arXiv:2305.13160}.

\bibitem[{Wang et~al.(2022)Wang, Wei, Schuurmans, Le, Chi, Narang, Chowdhery, and Zhou}]{wang2022self}
Xuezhi Wang, Jason Wei, Dale Schuurmans, Quoc Le, Ed~Chi, Sharan Narang, Aakanksha Chowdhery, and Denny Zhou. 2022.
\newblock Self-consistency improves chain of thought reasoning in language models.
\newblock \emph{arXiv preprint arXiv:2203.11171}.

\bibitem[{Wei et~al.(2022)Wei, Wang, Schuurmans, Bosma, Xia, Chi, Le, Zhou et~al.}]{wei2022chain}
Jason Wei, Xuezhi Wang, Dale Schuurmans, Maarten Bosma, Fei Xia, Ed~Chi, Quoc~V Le, Denny Zhou, and 1 others. 2022.
\newblock Chain-of-thought prompting elicits reasoning in large language models.
\newblock \emph{Advances in neural information processing systems}, 35:24824--24837.

\bibitem[{Xiong et~al.(2023{\natexlab{a}})Xiong, Deng, Koh, Wu, Li, Xu, and Hooi}]{xiong2023proximity}
Miao Xiong, Ailin Deng, Pang Wei~W Koh, Jiaying Wu, Shen Li, Jianqing Xu, and Bryan Hooi. 2023{\natexlab{a}}.
\newblock Proximity-informed calibration for deep neural networks.
\newblock \emph{Advances in Neural Information Processing Systems}, 36:68511--68538.

\bibitem[{Xiong et~al.(2023{\natexlab{b}})Xiong, Hu, Lu, Li, Fu, He, and Hooi}]{xiong2023can}
Miao Xiong, Zhiyuan Hu, Xinyang Lu, Yifei Li, Jie Fu, Junxian He, and Bryan Hooi. 2023{\natexlab{b}}.
\newblock Can llms express their uncertainty? an empirical evaluation of confidence elicitation in llms.
\newblock \emph{arXiv preprint arXiv:2306.13063}.

\bibitem[{Yang et~al.(2024)Yang, Tsai, and Yamada}]{yang2024verbalized}
Daniel Yang, Yao-Hung~Hubert Tsai, and Makoto Yamada. 2024.
\newblock On verbalized confidence scores for llms.
\newblock \emph{arXiv preprint arXiv:2412.14737}.

\bibitem[{Zadrozny and Elkan(2001)}]{zadrozny2001obtaining}
Bianca Zadrozny and Charles Elkan. 2001.
\newblock Obtaining calibrated probability estimates from decision trees and naive bayesian classifiers.
\newblock In \emph{Icml}, volume~1.

\bibitem[{Zhang et~al.(2020)Zhang, Kailkhura, and Han}]{zhang2020mix}
Jize Zhang, Bhavya Kailkhura, and T~Yong-Jin Han. 2020.
\newblock Mix-n-match: Ensemble and compositional methods for uncertainty calibration in deep learning.
\newblock In \emph{International conference on machine learning}, pages 11117--11128. PMLR.

\end{thebibliography}

\clearpage
\appendix

\section{Design Details about ConfMAD}
\label{sec:design}
\subsection{Prompt Design}
\label{sec:prompt}

In this section, we present the prompt design for ConfMAD. According to the ConfMAD framework, prompts are divided into two phases: the initial round (where LLM agents independently provide answers to queries) and the debate rounds (where LLM agents engage in sequential deliberation). When using LN confidence, confidence scores are computed based on token log probabilities. In contrast, SV confidence scores are elicited through prompts.

\subsubsection{Prompt design for LN confidence}

In the initial round, we only require the LLM agent to provide its reasoning (as arguments) and answer based on the question. Subsequently, we calculate the confidence score and append it to the reasoning and answer in the format "Confidence: [0-100]". In subsequent rounds, we concatenate the debate history to the User Prompt and instruct the LLM agent through the System Prompt to pay attention to the confidence scores of other debaters. If the format of the output from the LLM is disorganized and makes answer extraction difficult, consider slightly adjusting the prompts based on the specific dataset. For example, explicitly instructing the LLM agent to answer mathematical problems in LaTeX format. It is worth noting that prompting the model to generate reasoning first yields better results than requesting the answer first.

\begin{tcolorbox}[
  enhanced,
  colback=gray!10,
  colframe=gray!50!black,
  arc=5mm,
  boxrule=0.5mm,
  title={Init Round $r=0$},
  fonttitle=\bfseries\large,
  coltitle=black,
  colbacktitle=white,
  attach boxed title to top left={xshift=7mm, yshift=-2mm},
  boxed title style={boxrule=0.5pt, colframe=black} 
]
\textbf{System Prompt (Init):} \\
You are \{debater\}, a helpful AI assistant. You are now required to answer a question given by the user. 
Please give your reasoning for the question, and give your answer to the question. 
Please output in the following format strictly:

\noindent \textbf{Reason:} [your reason for the answer here]

\noindent \textbf{Answer:} [your answer here, only the answer]

\vspace{0.5em}
\textbf{User Prompt (Init):} \\
Question: \{question\}
\end{tcolorbox}

\begin{tcolorbox}[
  enhanced,
  colback=gray!10,
  colframe=gray!50!black,
  arc=5mm,
  boxrule=0.5mm,
  title={Debate Round $r>0$},
  fonttitle=\bfseries\large,
  coltitle=black,
  colbacktitle=white,
  attach boxed title to top left={xshift=7mm, yshift=-2mm},
  boxed title style={boxrule=0.5pt, colframe=black} 
]
\textbf{System Prompt (Debate):} \\
You are \{debater\}, a debater. You are now required to answer a question given by the user and debate with other debaters about the answer. PAY SPECIAL ATTENTION to these confidence scores as they reflect the reliability and conviction of each debater. If the confidence score is low, it may indicate uncertainty in the debater's answer. Please evaluate it further in this case. Then, formulate your debate arguments and provide your answer to the question.
Please output in the following format strictly:

\noindent \textbf{Reason:} [your debate arguments here]

\noindent \textbf{Answer:} [your answer here, only the answer]

\vspace{0.5em}
\textbf{User Prompt (Debate):} \\
Question: $\{$question$\}$ \\
Debate history: $\{$debate\_history$\}$
\end{tcolorbox}

\subsubsection{Prompt design for SV confidence}

Compared to the prompts in the LN confidence section, the prompt for SV confidence only requires the LLM agent to further generate a confidence score between 0-100.

\begin{tcolorbox}[
  enhanced,
  colback=gray!10,
  colframe=gray!50!black,
  arc=5mm,
  boxrule=0.5mm,
  title={Init Round $r=0$},
  fonttitle=\bfseries\large,
  coltitle=black,
  colbacktitle=white,
  attach boxed title to top left={xshift=7mm, yshift=-2mm},
  boxed title style={boxrule=0.5pt, colframe=black} 
]
\textbf{System Prompt (Init):} \\
You are $\{$debater$\}$, a helpful AI assistant. You are now required to answer a question given by the user. 
Please provide a clear reasoning for your answer, followed by your answer to the question. 
It is crucial to also include your confidence score, which reflects how strongly you believe your answer is correct. 
Consider the confidence score carefully as it represents the likelihood of your answer being accurate. 
Please output in the following format strictly:

\noindent \textbf{Reason:} [your reason for the answer here]

\noindent \textbf{Answer:} [your answer here, only the answer]

\noindent \textbf{Confidence score:} [your confidence score only, 0-100]

\vspace{0.5em}
\textbf{User Prompt (Init):} \\
Question: $\{$question$\}$
\end{tcolorbox}

\begin{tcolorbox}[
  enhanced,
  colback=gray!10,
  colframe=gray!50!black,
  arc=5mm,
  boxrule=0.5mm,
  title={Debate Round $r>0$},
  fonttitle=\bfseries\large,
  coltitle=black,
  colbacktitle=white,
  attach boxed title to top left={xshift=7mm, yshift=-2mm},
  boxed title style={boxrule=0.5pt, colframe=black} 
]
\textbf{System Prompt (Debate):} \\
You are $\{$debater$\}$, a debater. You are now required to answer a question given by the user and debate with other debaters about the answer. 
PAY SPECIAL ATTENTION to these confidence scores as they reflect the reliability and conviction of each debater's argument. 
If the confidence score is low, it may indicate uncertainty in the debater's answer. Please evaluate it further in this case.
Then, formulate your debate arguments and provide your answer to the question.
Finally, include your confidence score, which is a critical measure of how strongly you believe your answer is correct. 
Please output in the following format strictly:

\noindent \textbf{Reason:} [your debate arguments here]

\noindent \textbf{Answer:} [your answer here, only the answer]

\noindent \textbf{Confidence score:} [your confidence score only, 0-100]

\vspace{0.5em}
\textbf{User Prompt (Debate):} \\
Question: $\{$question$\}$ \\
Debate history: $\{$debate\_history$\}$
\end{tcolorbox}

For the No Conf prompt, we simply remove all expressions related to confidence from the LN prompt.

\subsection{Pseudo Code}
\label{sec:pesudo}

Algorithm \ref{alg:conf_debate} presents the pseudocode design of the ConfMAD system. For the No Conf setting, simply remove all content related to confidence expressions in Algorithm \ref{alg:conf_debate}.

\begin{algorithm}
\small
\caption{ConfMAD}
\label{alg:conf_debate}
\SetAlgoLined
\KwIn{Models $M_1,\dots,M_n$, Question $x$, Prompt $p$, Max Debate Round $T$}
\KwOut{Final answer $A_{\text{f}}$}
\For{$i \leftarrow 1$ \KwTo $n$}{
    $R_i^0, A_i^0, C_i^{\prime 0} \leftarrow M_i(x \mid p)$ \\
    $C_i^0 \leftarrow \text{Calibration}(C_i^{\prime 0})$ \\
    $D_i^0 \leftarrow \langle R_i^0, A_i^0, C_i^0 \rangle$
}
$H_{0} \leftarrow \text{Concat}(D_1^{0},\dots,D_n^{0})$ \\
\For{$r \leftarrow 1$ \KwTo $T$}{
    $\hat{H}_{r} \leftarrow H_{r-1}$ \\
    \For{$i \leftarrow 1$ \KwTo $n$}{
        $R_i^r, A_i^r, C_i^{\prime r} \leftarrow M_i(x \mid p, \hat{H}_r)$ \\
        $C_i^r \leftarrow \text{Calibration}(C_i^{\prime r})$ \\
        $D_i^r \leftarrow \langle R_i^r, A_i^r, C_i^r \rangle$ \\
        $\hat{H}_r \leftarrow \text{Concat}(\hat{H}_r, D_i^r)$
    }
    $H_{r} \leftarrow \hat{H}_{r}$
}
\Return{$A_{\text{f}} \leftarrow A_{i^*}^{T}$, where $i^* = \arg\max_{i} C_i^{T}$}
\end{algorithm}

\subsection{Datasets}
\label{sec:size}

The sizes of the training and validation sets for each dataset are shown in Table \ref{tab:Size}. The BIGGSM dataset is publicly accessible on GitHub, serving to further evaluate LLMs' reasoning capabilities on complex mathematical problems. The usage of BBH, MMLU, and MATH datasets all complies with the MIT License. BBH is designed to test general reasoning across diverse tasks, MMLU evaluates knowledge across multiple academic subjects, and MATH focuses on high-school competition-level math problems. Our use of these datasets aligns with their original intended purposes for benchmarking LLMs.

\begin{table}[ht]
  \centering
  \resizebox{\columnwidth}{!}{
    \begin{tabular}{lcccc}
      \toprule
      \textbf{Dataset} & \textbf{BIGGSM} & \textbf{BBH} & \textbf{MMLU} & \textbf{MATH} \\
      \midrule
      Test   & 400 & 2,000 & 2,000 & 1,000 \\
      Valid. & 200 & 1,000 & 1,000 & 1,000 \\
      \bottomrule
    \end{tabular}
  }
  \caption{The sizes of the test and validation sets(Valid.) from different benchmarks. Validation sets are used to train calibration models.}
  \label{tab:Size}
\end{table}

\section{Extended Results}
\label{sec:extended}
\subsection{Calibration Metric of ConfMAD}
 
\textit{Expected Calibration Error} (ECE) is a commonly used metric for evaluating the quality of confidence calibration. 
It quantifies the gap between predicted confidence and empirical accuracy across different confidence intervals and is defined as:

\begin{equation}
\label{eq:ece}
\text{ECE} = \sum_{m=1}^{M} \frac{|B_m|}{n} \left| \text{acc}(B_m) - \text{conf}(B_m) \right|
\end{equation}

\begin{figure*}[ht]
  \centering
  \includegraphics[width=\textwidth]{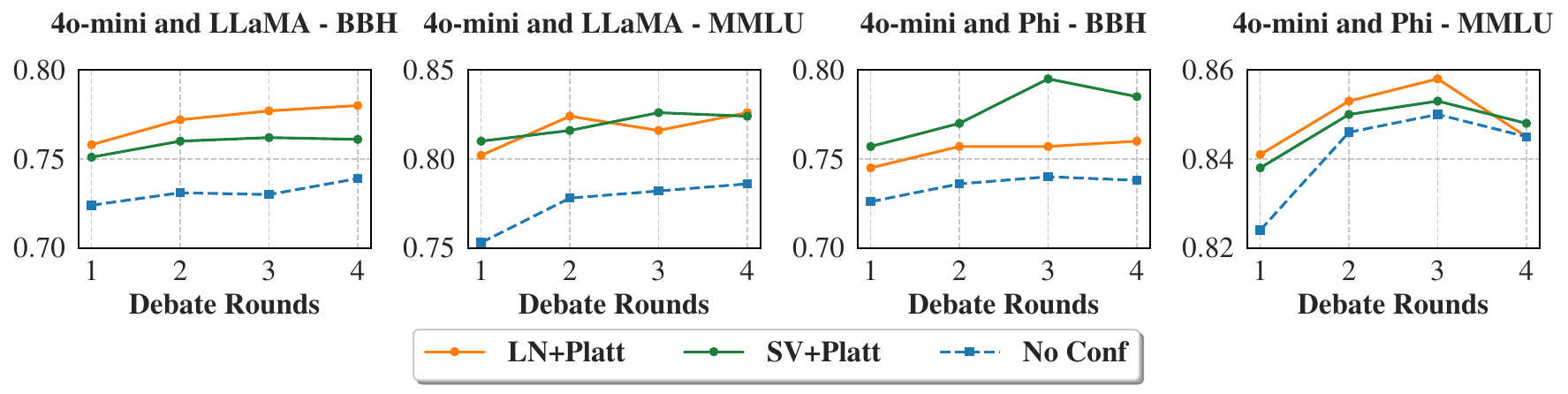}
  \caption{Final accuracy over debate rounds ($1\leq r \leq4$) for 500 randomly sampled questions from MMLU and BBH datasets, comparing 4o-mini debating with LLaMA and Phi.}
  \label{fig:debate_round}
\end{figure*}

\begin{table*}[htbp]
  \centering
  % \resizebox{\linewidth}{!}{
    \begin{tabular}{l|l|cccccc}
      \toprule
      \multirow{2}{*}{\textbf{Model}} & \multirow{2}{*}{\textbf{Dataset}} 
      & \multicolumn{4}{c}{\textbf{LN Confidence}} 
      & \multicolumn{2}{c}{\textbf{SV Confidence}} \\
      & & Platt & Temp & Vanilla &  & Platt & Vanilla \\
      \midrule
      \multirow{4}{*}{4o-mini} 
      & MMLU   & 0.2 & 2.6 & 19.4 & & 8.0 & 13.8 \\
      & BBH    & 1.7 & 4.2 & 22.4 & & 1.8 & 19.3 \\
      & BIGGSM & 3.2 & 5.1 & 53.0 & & 2.1 & 57.4 \\
      & MATH   & 2.9 & 7.2 & 24.8 & & 8.8 & 28.8 \\
      \midrule
      \multirow{4}{*}{LLaMA} 
      & MMLU   & 4.7 & 17.9 & 8.8 & & 1.8 & 15.2 \\
      & BBH    & 4.8 & 26.6 & 24.7 & & 3.3 & 25.3 \\
      & BIGGSM & 8.7 & 15.2 & 50.4 & & 4.0 & 68.9 \\
      & MATH   & 4.2 & 12.0 & 30.9 & & 0.5 & 25.2 \\
      \midrule
      \multirow{4}{*}{Phi} 
      & MMLU   & 1.4 & 2.2 & 15.3 & & 5.0 & 15.4 \\
      & BBH    & 6.0 & 8.5 & 25.2 & & 4.0 & 26.3 \\
      & BIGGSM & 0.7 & 5.0 & 23.0 & & 3.0 & 26.6 \\
      & MATH   & 0.3 & 4.4 & 22.7 & & 8.8 & 28.8 \\
      \bottomrule
    \end{tabular}
  % }
    \caption{Comparison of ECE $\Downarrow$ between original and calibrated confidence scores on validation sets. Validation sets are used to train calibration models. Results are given $\times 100$.}

  \label{tab:ece_compare_train}
\end{table*}

In Equation \ref{eq:ece}, \( M \) denotes the number of confidence bins, and \( B_m \) represents the set of samples whose confidence scores fall into the \( m \)-th bin. The total number of samples is denoted by \( n \). The term \( \text{acc}(B_m) \) refers to the empirical accuracy within bin \( B_m \), while \( \text{conf}(B_m) \) denotes the average predicted confidence in the same bin. A lower ECE indicates a better-calibrated result. In our experiments, we set \(M=10\) when computing ECE scores.

We first present the ECE scores of different calibration methods on the validation set. Table~\ref{tab:ece_compare_train} presents the ECE scores obtained with LN and SV when applying Platt Scaling, Temperature Scaling, and when no calibration is applied. The training process of Histogram Binning is highly consistent with the way ECE is computed; its ECE scores on the training set are always close to zero. Therefore, we do not present them in the table.

Table~\ref{tab:ece_compare_test} compares the calibration metrics of the final debate round with and without applying calibration methods. It can be observed that Platt Scaling is generally a more stable calibration method under both LN and SV confidence settings, and it consistently leads to improvements in ECE. In contrast, Histogram Binning and Temperature Scaling show less stability and, in some cases, may even result in increased ECE.

\begin{table*}[htbp]
  \centering
  % \resizebox{\linewidth}{!}{
    \begin{tabular}{l|l|ccccccc}
      \toprule
      \multirow{2}{*}{\textbf{Partner}} & \multirow{2}{*}{\textbf{Task}} 
      & \multicolumn{4}{c}{\textbf{LN Confidence}} 
      & \multicolumn{3}{c}{\textbf{SV Confidence}} \\
      & & Platt & Histo & Temp & Vanilla & Platt & Histo & Vanilla \\
      \midrule
      \multirow{4}{*}{LLaMA} 
      & BIGGSM & 7.1 & 12.3 & 30.6 & 37.2 & 22.6 & 36.0 & 37.0 \\
      & BBH    & 0.6 & 2.8  & 14.5 & 24.7 & 11.7 & 26.6 & 21.7 \\
      & MMLU   & 0.5 & 5.3  & 5.1  & 19.4 & 4.8  & 2.2  & 13.6 \\
      & MATH   & 4.5 & 23.1 & 4.2  & 39.1 & 10.7 & 27.2 & 26.0 \\
      \midrule
      \multirow{4}{*}{Phi} 
      & BIGGSM & 6.8 & 9.1  & 8.2  & 31.5 & 0.8  & 31.4 & 30.4 \\
      & BBH    & 1.2 & 2.9  & 10.9 & 21.9 & 9.2  & 26.4 & 21.8 \\
      & MMLU   & 1.0 & 1.8  & 6.5  & 18.2 & 10.2 & 11.8 & 14.3 \\
      & MATH   & 2.5 & 9.6  & 6.2  & 21.7 & 6.2  & 24.1 & 21.0 \\
      \bottomrule
    \end{tabular}
  % }
  \caption{Comparison of ECE $\Downarrow$ with and without calibration across different datasets using ConfMAD debates of the final results. Results are given $\times 100$.}
  \label{tab:ece_compare_test}
\end{table*}

However, it is important to note that the performance of a calibration method depends not only on its design, but also on the quality and size of the validation set, as well as the quality of the confidence elicited from LLMs on that set. Therefore, improving the calibration of ConfMAD results requires consideration from multiple perspectives, including ensuring that the validation set is sufficiently large and diverse in confidence scores.

\begin{figure}[t]
  \includegraphics[width=\columnwidth]{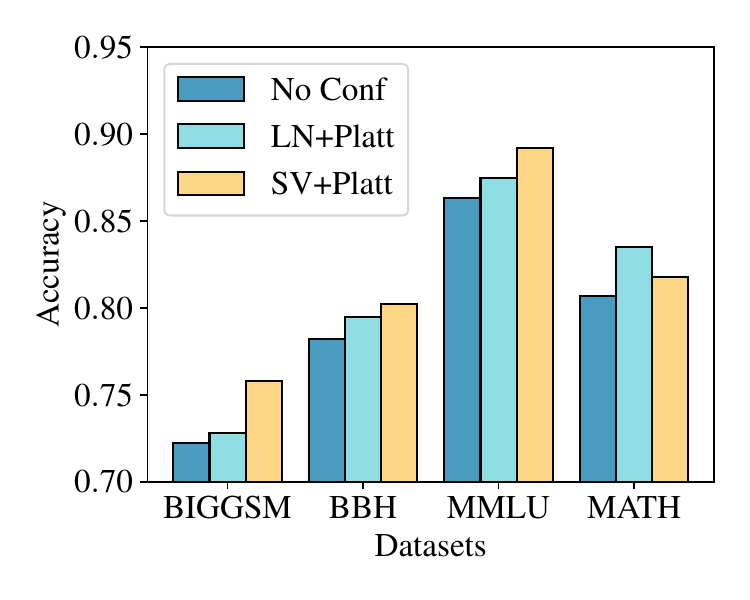}
  \caption{Comparison of debate accuracy with and without confidence scores across four datasets. The selected LLMs are 4o-mini, LLaMA, and Phi, with three debate rounds (one initial round followed by two one-by-one rounds).}
  \label{fig:ThreeAgent}
\end{figure}

\subsection{Rounds of Debate}
\label{sec:round}

Figure \ref{fig:debate_round} shows the relationship between debate rounds and the accuracy of MAD systems, based on subsets of 500 randomly sampled questions from MMLU and BBH. Accuracy generally peaks within the first 2–3 debate rounds and remains relatively stable across most settings. In some cases, increasing debate rounds beyond this point can even degrade the system's performance. These observations are consistent with findings from prior related work \citep{du2023improving, estornell2024multi, liu2024groupdebate}. It is important to note that, given the multiple inferences and longer context in MAD systems, choosing an appropriate number of debate rounds helps reduce computational overhead.

\subsection{Number of Agents}
\label{sec:agents}

We conducted debates involving three LLM agents: 4o-mini, LLaMA, and Phi. These debates were performed under three settings: No Conf, LN+Platt, and SV+Platt. There are four datasets, each comprising 400 samples: BIGGSM, and randomly selected subsets from MMLU, BBH, and MATH. Each debate spanned three rounds. The outcomes of this experiment are presented in Figure \ref{fig:ThreeAgent}. In this three-agent setting, we observed that debating with confidence continued to enhance final accuracy compared with the No Conf setting. However, the magnitude of improvement was less pronounced compared to previous two-agent settings. 

\subsection{Correction Cases}
\label{sec:correction}
Table \ref{tab:Correction} presents the number of correction cases under different debate settings.

\subsection{Categorical Confidence}
\label{sec:categorical_confidence}

Considering that LLMs are essentially generative models, using a fine-grained confidence scale from 0 to 100 may be overly precise, and LLMs may struggle to accurately express their confidence levels. We conducted experiments comparing coarse-grained confidence (Categorical Confidence) with fine-grained confidence (Raw Confidence). In the debate setting, we scaled the confidence expression by dividing it by 10 and then rounding the result, and compared it against the original range. We evaluated two debaters, 4o-mini and Phi, using Platt Scaling for calibration. As shown in Table~\ref{tab:granularity}, the performance difference between LN and SV is negligible on MMLU and MATH. However, on BBH and BIGGSM, Categorical Confidence exhibits noticeable fluctuations. We hypothesize that this is because overly coarse confidence expressions undermine the effectiveness of the calibration method.

In addition to expressing confidence levels in numerical form, it is also possible to represent them using textual categories, such as \textit{high}, \textit{medium}, and \textit{low}. Future work could explore how to integrate such textual confidence expressions with calibration methods.

\begin{table}[ht]
  \centering
  \resizebox{\columnwidth}{!}{
    \begin{tabular}{l|l|cccc}
      \toprule
      Debaters & Setting & BIGGSM & BBH & MMLU & MATH \\
      \midrule
      \multirow{6}{*}{\makecell{4o-mini \\ LLaMA}}
      & No Conf       & 129 & 293 & 131 & 182 \\
      & LN+Platt      & 132 & 337 & 167 & 181 \\
      & LN+Histo      & 133 & 271 & 203 & 205 \\
      & LN+Temp       & 120 & 343 & 219 & 212 \\
      & SV+Platt      & 116 & 289 & 164 & 215 \\
      & SV+Histo      & 106 & 295 & 131 & 235 \\
      \midrule
      \multirow{6}{*}{\makecell{4o-mini \\ Phi}}
      & No Conf       & 35 & 244 & 165 & 136 \\
      & LN+Platt      & 72 & 298 & 223 & 160 \\
      & LN+Histo      & 61 & 296 & 190 & 158 \\
      & LN+Temp       & 82 & 293 & 186 & 164 \\
      & SV+Platt      & 56 & 299 & 196 & 173 \\
      & SV+Histo      & 36 & 263 & 193 & 183 \\
      \bottomrule
    \end{tabular}
  }
  \caption{Number of correction cases under different debate settings. Each value indicates how often an initially incorrect answer was corrected during the debate.}
  \label{tab:Correction}
\end{table}

\begin{table}[ht]
  \centering
  \resizebox{\linewidth}{!}{
    \begin{tabular}{l|l|cccc}
      \toprule
      Granularity & Conf. & BIGGSM & BBH & MMLU & MATH \\
      \midrule
      \multirow{2}{*}{Categorical} 
      & LN & 0.713 & 0.774 & 0.831 & 0.789 \\
      & SV & 0.740 & 0.756 & 0.824 & 0.777 \\
      \midrule
      \multirow{2}{*}{Raw Conf.} 
      & LN & 0.747 & 0.777 & 0.835 & 0.785 \\
      & SV & 0.730 & 0.781 & 0.834 & 0.784 \\
      \bottomrule
    \end{tabular}
  }
  \caption{Performance comparison between coarse-grained (Categorical, 0–10) and fine-grained (Raw Conf.) confidence expression. Debaters are 4o-mini and Phi with Platt scaling calibration.}
  \label{tab:granularity}
\end{table}

\begin{table*}[ht]
  \centering
    \begin{tabular}{l|l|l|cccc}
      \toprule
      Debaters & Calibration & Mode & BIGGSM & BBH & MMLU & MATH \\
      \midrule
      \multirow{4}{*}{\makecell{4o-mini \\ LLaMA}}
      & \multirow{2}{*}{Platt+LN} 
          & One-by-One & 0.632 & 0.763 & 0.833 & 0.711 \\
      &   & Broadcast  & 0.632 & 0.755 & 0.825 & 0.723 \\
      \cmidrule(l){2-7}
      & \multirow{2}{*}{Platt+SV} 
          & One-by-One & 0.655 & 0.767 & 0.831 & 0.725 \\
      &   & Broadcast  & 0.642 & 0.753 & 0.824 & 0.711 \\
      \midrule
      \multirow{4}{*}{\makecell{4o-mini \\ Phi}}
      & \multirow{2}{*}{Platt+LN} 
          & One-by-One & 0.747 & 0.777 & 0.835 & 0.785 \\
      &   & Broadcast  & 0.735 & 0.768 & 0.834 & 0.792 \\
      \cmidrule(l){2-7}
      & \multirow{2}{*}{Platt+SV} 
          & One-by-One & 0.730 & 0.781 & 0.834 & 0.784 \\
      &   & Broadcast  & 0.733 & 0.775 & 0.843 & 0.784 \\
      \bottomrule
    \end{tabular}
  \caption{Performance comparison between One-by-One and Broadcast debate modes under Platt Scaling. Confidence elicition methods are LN and SV.}
  \label{tab:debate_modes}
\end{table*}

\begin{table*}[!htbp]
  \centering
  \resizebox{\textwidth}{!}{
    \begin{tabular}{ll|ccc|ccc}
      \toprule
      \textbf{Dataset} & \textbf{Setting} 
      & \multicolumn{3}{c|}{\textbf{4o-mini + LLaMA}} 
      & \multicolumn{3}{c}{\textbf{4o-mini + Phi}} \\
      
      & 
      & WR of Round 0 & WR of Round 1 & Acc. 
      & WR of Round 0 & WR of Round 1 & Acc. \\
      \midrule

      \multirow{7}{*}{BIGGSM}
        & LN+Platt   & 0.796 (129/162) & 0.435 (10/23) & 0.632 & 0.816 (84/103) & 0.880 (22/25) & 0.747 \\
        & LN+Histo   & 0.865 (134/155) & 0.583 (7/12)  & 0.627 & 0.848 (78/92) & 0.842 (16/19) & 0.735 \\
        & LN+Temp    & 0.895 (136/152) & 0.773 (17/22) & 0.625 & 0.875 (91/104) & 0.938 (15/16) & 0.748 \\
        & LN+Vanilla & 0.887 (125/141) & 0.500 (7/14)  & 0.627 & 0.224 (22/98) & 0.167 (4/24) & 0.694 \\
        & SV+Platt   & 0.852 (127/149) & 0.481 (13/27) & 0.655 & 0.831 (69/83) & 0.806 (25/31) & 0.730 \\
        & SV+Histo   & 0.763 (106/139) & 0.533 (8/15)  & 0.620 & \cellcolor{gray!20}0.268 (22/82) & \cellcolor{gray!20}0.000 (0/26) & \cellcolor{gray!20}0.675 \\
        & SV+Vanilla & \cellcolor{gray!20}0.577 (41/71) & \cellcolor{gray!20}0.667 (12/18)   & \cellcolor{gray!20}0.608 & 0.506 (42/83) & 0.542 (13/24) & 0.708 \\
      \midrule

      \multirow{7}{*}{BBH}
        & LN+Platt   & 0.645 (335/519) & 0.534 (124/232) & 0.763 & 0.614 (316/515) & 0.514 (111/216) & 0.777 \\
        & LN+Histo   & 0.534 (268/502) & 0.565 (105/186) & 0.753 & \cellcolor{gray!20}0.552 (280/507) & \cellcolor{gray!20}0.488 (81/166) & \cellcolor{gray!20}0.753 \\
        & LN+Temp    & 0.632 (350/554) & 0.509 (115/226) & 0.751 & 0.668 (334/500) & 0.550 (120/218) & 0.780 \\
        & LN+Vanilla & \cellcolor{gray!20} 0.664 (334/503) & \cellcolor{gray!20} 0.354 (80/226)  & \cellcolor{gray!20} 0.748 & 0.551 (297/539) & 0.464 (96/207) & 0.778 \\
        & SV+Platt   & 0.648 (309/477) & 0.530 (149/281) & 0.767 & 0.576 (297/516) & 0.489 (116/237) & 0.778 \\
        & SV+Histo   & 0.534 (275/515) &  0.524 (86/164) & 0.757 & 0.526 (253/481) & 0.533 (88/165) & 0.757 \\
        & SV+Vanilla & 0.625 (223/357) & 0.682 (122/179) & 0.763 & 0.408 (194/476) & 0.452 (94/208) & 0.760 \\
      \midrule

      \multirow{7}{*}{MMLU}
        & LN+Platt   & 0.623 (230/369) & 0.566 (142/251) & 0.833 & 0.668 (263/394) & 0.575 (100/174) & 0.835 \\
        & LN+Histo   & 0.674 (256/380) & 0.592 (161/272) & 0.819 & 0.676 (244/361) & 0.644 (112/174) & 0.832 \\
        & LN+Temp    & 0.697 (274/393) & 0.554 (107/193) & 0.829 & 0.683 (231/338) & 0.585 (86/147) & 0.833 \\
        & LN+Vanilla & 0.441 (154/349) & 0.376 (77/205)  & 0.804 & \cellcolor{gray!20}0.608 (129/212) & \cellcolor{gray!20}0.273 (27/99) & \cellcolor{gray!20}0.801 \\
        & SV+Platt   & 0.694 (249/359) & 0.594 (148/249) & 0.829 & 0.663 (216/326) & 0.557 (83/149) & 0.834 \\
        & SV+Histo   & \cellcolor{gray!20}0.499 (178/357) & \cellcolor{gray!20}0.529 (118/223) & \cellcolor{gray!20}0.805 & 0.529 (189/357) & 0.548 (91/166) & 0.815 \\
        & SV+Vanilla    & 0.475 (172/362) & 0.660 (192/291) & 0.824 & 0.669 (206/308) & 0.624 (93/149) & 0.829 \\
      \midrule

      \multirow{7}{*}{MATH}
        & LN+Platt   & 0.815 (190/233) & 0.681 (32/47) & 0.719 & 0.608 (119/194) & 0.766 (36/47) & 0.788 \\
        & LN+Histo   & \cellcolor{gray!20}0.772 (190/246) & \cellcolor{gray!20}0.641 (25/39) & \cellcolor{gray!20}0.711 & 0.576 (121/210) & 0.370 (17/46) & 0.788 \\
        & LN+Temp & \cellcolor{gray!20}0.782 (190/243) & \cellcolor{gray!20}0.676 (23/34) & \cellcolor{gray!20}0.711 & \cellcolor{gray!20}0.631 (128/203) & \cellcolor{gray!20}0.556 (25/45) & \cellcolor{gray!20}0.765 \\
        & LN+Vanilla & 0.815 (203/249) & 0.627 (32/51) & 0.719 & 0.562 (109/194) & 0.311 (14/45) & 0.781 \\
        & SV+Platt   & 0.716 (174/243) & 0.460 (57/124) & 0.715 & 0.774 (154/199) & 0.733 (44/60) & 0.784 \\
        & SV+Histo   & 0.696 (176/253) & 0.504 (57/113) & 0.720 &  0.721 (158/219) & 0.322 (19/59) & 0.784 \\
        & SV+Vanilla & 0.421 (61/145) & 0.519 (27/52) & 0.730 & 0.424 (92/217) & 0.507 (34/67) & 0.784 \\

      \bottomrule
    \end{tabular}
  }
  \caption{
    Win Rate (WR) and accuracy under different confidence settings across datasets. Metrics are reported for two model pairs: 4o-mini + LLaMA and 4o-mini + Phi. Gray cells indicate the worst-performing setting for each dataset and model pair.
  }
  \label{tab:winrate_compact}
\end{table*}

\subsection{Win Rate Analysis}
\label{sec:win_rate_analysis}

Table~\ref{tab:winrate_compact} presents the Win Rates (WR) across all debate settings. We observe that, under the same confidence expression method, lower WR values are usually associated with poorer performance. This suggests that, when LLMs disagree, the relative ordering of confidence scores plays a crucial role in influencing the final outcome. However, in some cases, a lower WR does not necessarily lead to a drop in performance. For example, on the MATH dataset, even though uncalibrated settings tend to have lower WRs, the overall accuracy does not degrade. We attribute this to a higher frequency of tie confidence scores, where neither LLM agent clearly dominates. In addition, we notice an interesting phenomenon when using 4o-mini and LLaMA as debating agents. On the MATH dataset, although LLaMA initially has lower accuracy, it achieves stronger performance after the debate. This may suggest that confidence debate could have encouraged more reflection and adjustment during the debate process. These findings further highlight that the role of confidence expression in MAD systems is nuanced and complex.

\subsection{One-by-One Debate Mode v.s. Broadcast Debate Mode}
\label{sec:onebyone_broadcast}

Table~\ref{tab:debate_modes} details the performance comparison between the one-by-one and broadcast debate modes under Platt Scaling across various datasets. In the majority of cases, no significant performance disparity is observed between the two modes. A key feature of the broadcast mode is that it allows debaters to present arguments in parallel within each debate round, obviating the need to await the outputs of others and thereby accelerating the whole debate process. We adopted the one-by-one mode in this study primarily to maintain alignment with the MAD framework of \citet{du2023improving}, which facilitates a more direct performance comparison.

\subsection{Case Study}
\label{sec:case}

\begin{figure*}[ht]
  \centering
  \includegraphics[width=\textwidth]{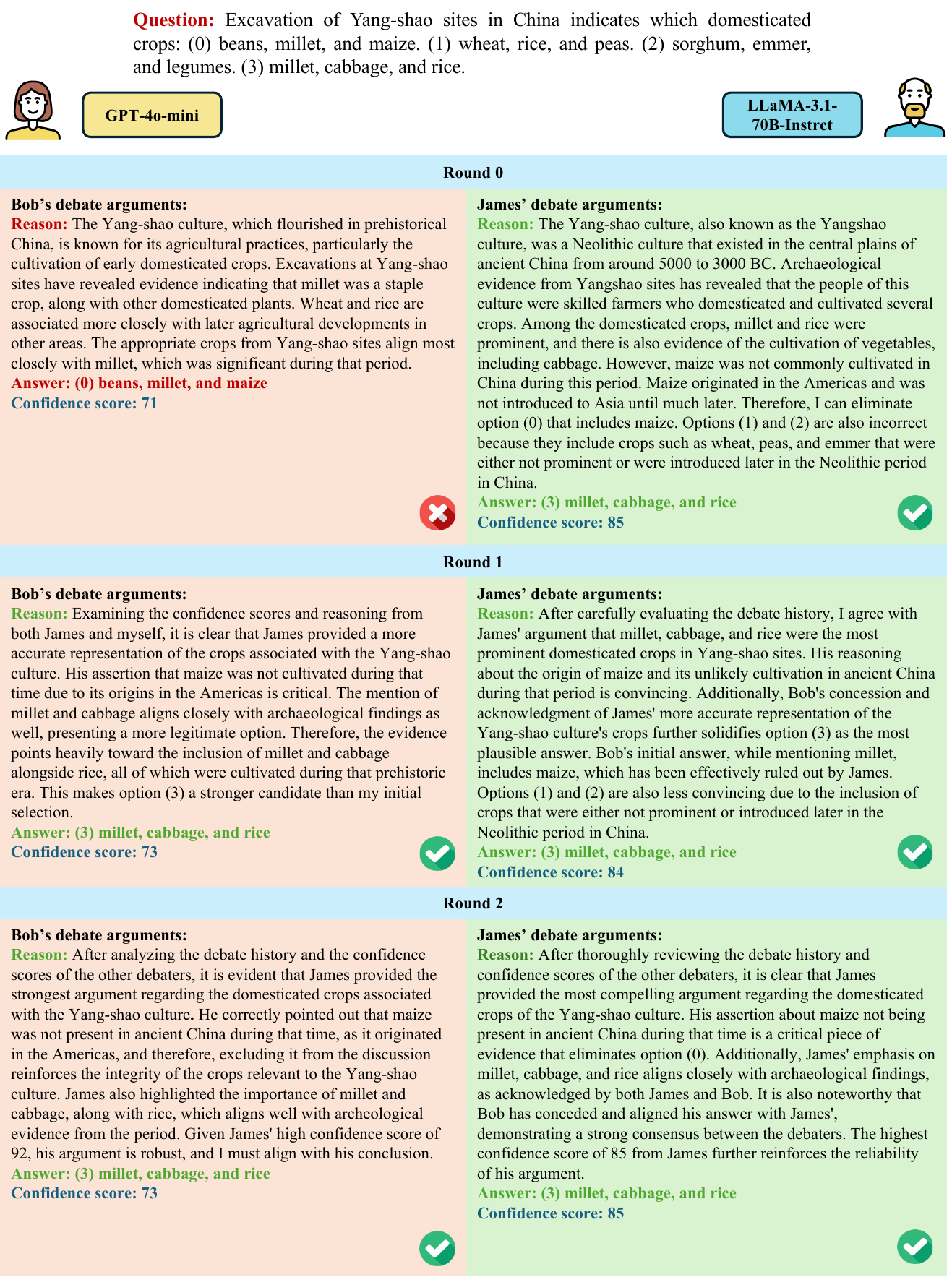}
  \caption{In a ConfMAD debate involving 4o-mini and LLaMA on a MMLU question under the LN+Platt setting, 4o-mini referenced LLaMA's confidence scores and answers, leading the debate to converge on the correct result.}
  \label{fig:Case1}
\end{figure*}

\begin{figure*}[ht]
  \centering
  \includegraphics[width=\textwidth]{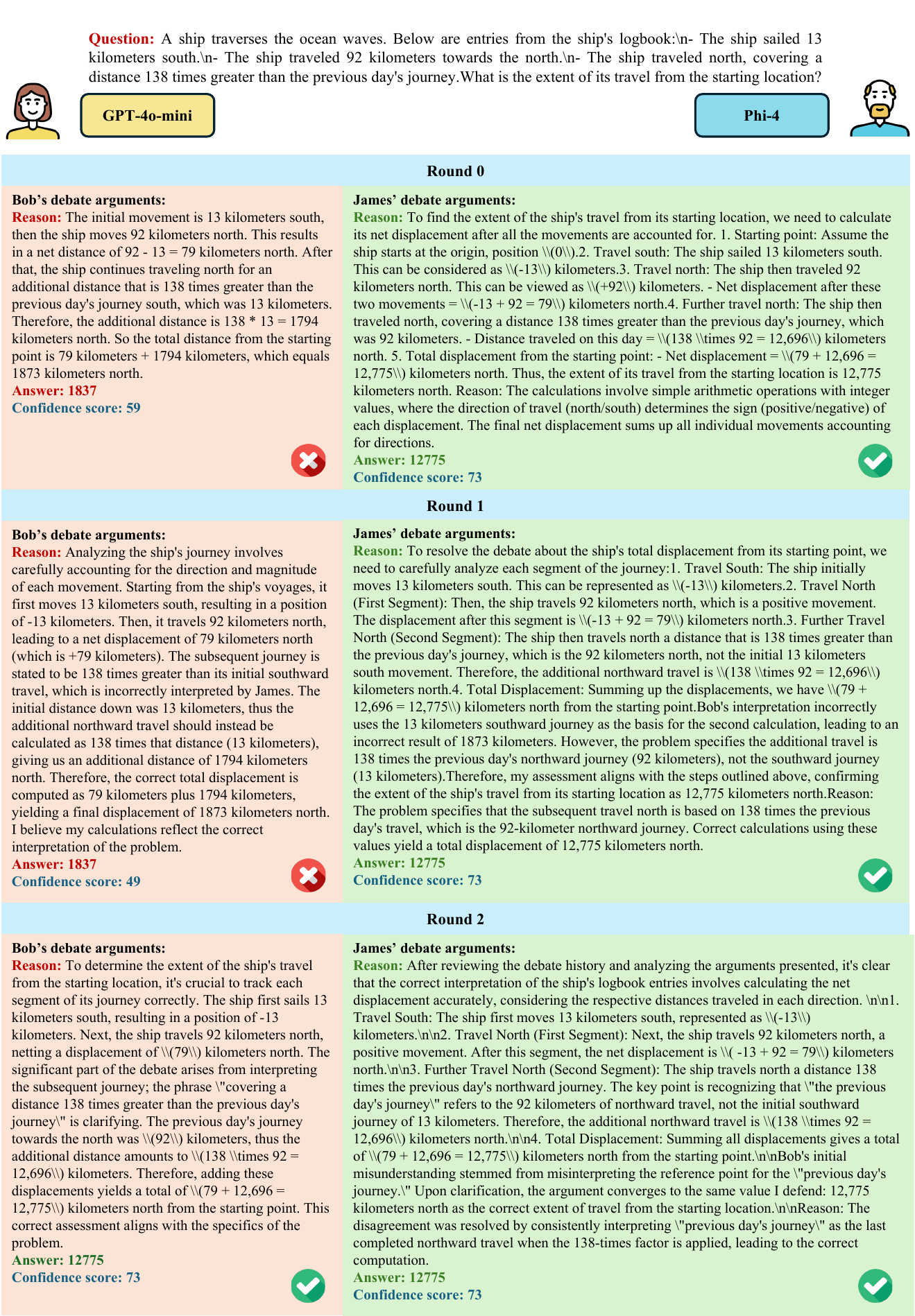}
  \caption{In a ConfMAD debate involving 4o-mini and Phi on a MATH question under the SV+Platt setting, 4o-mini referenced Phi's confidence scores and answers, leading the debate to converge on the correct result.}
  \label{fig:Case2}
\end{figure*}

In this section, we present several cases where debates conducted using the ConfMAD framework successfully converged to correct outcomes. Figure \ref{fig:Case1} and Figure \ref{fig:Case2} illustrate debate examples involving the 4o-mini and LLaMA models, as well as the Phi model, respectively.

\end{document}